\def\year{2022}\relax
\documentclass[letterpaper]{article} 
\usepackage{aaai22}  
\usepackage{times}  
\usepackage{helvet}  
\usepackage{courier}  
\usepackage[hyphens]{url}  
\usepackage{graphicx} 
\usepackage{natbib}  
\usepackage{caption} 
\DeclareCaptionStyle{ruled}{labelfont=normalfont,labelsep=colon,strut=off} 
\usepackage{algorithm}
\usepackage{algorithmic}
\usepackage{bbm}
\usepackage{newfloat}
\usepackage{listings}
\floatstyle{ruled}
\newfloat{listing}{tb}{lst}{}
\floatname{listing}{Listing}

\usepackage{amsmath,amssymb,amsthm,mathtools,bm}
\usepackage{kotex} 
\usepackage{nccmath}
\usepackage{color}

\usepackage{array,multirow}
\usepackage{caption}
\usepackage{tabularx}
\usepackage{booktabs}
\usepackage{wrapfig}
\usepackage{adjustbox}

\usepackage{arydshln}
\usepackage{caption}

\newcommand{\BF}[1]{\textbf{#1}}
\DeclareMathOperator*{\minimize}{min.}
\definecolor{mygray}{gray}{0.6}
\newcommand{\LineComment}[1]{\textcolor{mygray}{/*{ #1 }*/}}
\newcommand{\COMMENTmy}[1]{\hfill \textcolor{mygray}{// {#1}}}

\title{Graph Transplant: Node Saliency-Guided Graph Mixup  \\ with Local Structure Preservation}
\author{
    Joonhyung Park\equalcontrib, \textsuperscript{\rm 1}
    Hajin Shim\equalcontrib, \textsuperscript{\rm 1}
    Eunho Yang \textsuperscript{\rm 1,2}
}
\affiliations{
    \textsuperscript{\rm 1}Korea Advanced Institute of Science and Technology (KAIST) \\
    \textsuperscript{\rm 2}AITRICS \\

    \{deepjoon, shimazing, eunhoy\}@kaist.ac.kr
}

\usepackage{bibentry}

\begin{document} 

\maketitle

\begin{abstract}
Graph-structured datasets usually have irregular graph sizes and connectivities, rendering the use of recent data augmentation techniques, such as Mixup, difficult. To tackle this challenge, we present the first Mixup-like graph augmentation method at the graph-level called \textit{Graph Transplant}, which mixes irregular graphs in data space. To be well defined on various scales of the graph, our method identifies the sub-structure as a mix unit that can preserve the local information. Since the mixup-based methods without special consideration of the context are prone to generate noisy samples, our method explicitly employs the node saliency information to select meaningful subgraphs and adaptively determine the labels. We extensively validate our method with diverse GNN architectures on multiple graph classification benchmark datasets from a wide range of graph domains of different sizes.   
Experimental results show the consistent superiority of our method over other basic data augmentation baselines. 
We also demonstrate that \emph{Graph Transplant} enhances the performance in terms of robustness and model calibration.

\end{abstract}

\section{Introduction}
\label{intro}
Graph classification plays a critical role in a variety of fields from chemistry to social data analysis. In recent years, attempts to use graph neural networks (GNNs) to solve graph classification tasks have been in the spotlight \cite{neural_message_passing, ecc, diffpool, fair, isonn}
, as deep models successfully learn from unstructured data in various domains such as computer vision (CV) \cite{resnet} or natural language processing (NLP) \cite{transformer}. However, the challenge is that, like most deep models, in order for GNNs to successfully generalize even for unseen graph data, we are often required 
so much data that it is not practical. 

To train these data-hungry neural networks more cost-efficiently, data augmentation techniques are commonly employed. 
Existing graph augmentation methods largely rely on injecting structural noise such as edge manipulation \cite{rong2019dropedge, augforgraphclf} or node deletion \cite{graphcl} without considering the semantics of each graph data so that they can mislead the network due to their context-agnostic nature. 
Furthermore, most of the proposed graph augmentation strategies focus on node classification tasks \cite{graphmix, rong2019dropedge}, which is difficult to apply directly to graph classification.


As another attractive option, we consider Mixup \cite{mixup} which has become one of the standard data augmentations 
with its excellent performance in image classification \cite{cutmix}, sentence classification \cite{nlpmixup}, model robustness \cite{augmix}, etc. However, since existing Mixup variants \cite{manifold_mixup, cutmix, 
adamixup, comixup} are somewhat specific to image data or require data instances to lie in the same dimensional space, 
they are inappropriate for the case of graph data where there is usually no node correspondence, and even the number of nodes may differ across graph instances. 

In this paper, we propose \emph{Graph Transplant}, a novel Mixup-based data augmentation for graph classification tasks that can be generally used for multiple-domain datasets regardless of graph geometries.
To make sense of mixing graph instances with no node correspondence and the different number of nodes, \emph{Graph Transplant} selects a subgraph from each of the destination and source instances and transplants the subgraph of the source onto the destination whose subgraph is removed.
However, transplanting a subgraph from one instance to another may not properly align the final product with the corresponding interpolated label if the original charge of the subgraph is not reflected in the label.  
To address this issue, we propose to utilize the node saliency for selecting subgraphs and assigning labels. Specifically for our purpose, we define the node saliency as $\ell_2$ norm of the gradient of classification loss with respect to the output of the graph convolutional layer. 
It can be regarded as a measure of node's contribution when the model classifies the graph.
Furthermore, the label for the mixed graph is determined as the ratio of the saliency between the source and destination subgraphs. Instead of na\"ively using the ratio of the number of nodes in subgraphs, saliency-based label mixing describes the mixed graphs properly by weighting nodes according to their importance. Another non-trivial challenge in mixing graphs is to determine the connectivity between subgraphs obtained from two different instances. To this end, we suggest the two variants: 1) random connection under the constraint on the original node degree and 2) edge prediction based on node features. 

In summary, our contribution is threefold:
\begin{itemize}
    \item We propose \emph{Graph Transplant}, the first input-level Mixup-based graph augmentation, that can mix two dissimilar-structured graphs by replacing the destination subgraph with the source subgraph 
    while preserving the local structure.
    \item In contrast to random augmentations which risk producing noisy samples, we present a novel strategy to generate properly mixed graphs with adaptively assigned labels that adheres to the corresponding labels by explicitly leveraging the node saliency.
    \item Through extensive experiments, we show that our \emph{Graph Transplant} is an effective algorithm that can bring overall improvement across the multiple fields of graphs analysis for diverse architectures in terms of the classification performance, robustness, and model calibration. 
\end{itemize}

\section{Related Works}



\paragraph{Data augmentation for graph-structured data}

Most of the data augmentation methods for graph-structured data are biased to node classification tasks. In addition, many of them largely rely on edge perturbation. GraphSage \cite{hamilton2017inductive} samples a different subset of neighbors uniformly for every iteration. DropEdge \cite{rong2019dropedge} randomly removes a fraction of edges before each training epoch. AdaEdge \cite{adaedge} iteratively adds or removes edges between nodes that are predicted to have the same labels with high confidence. \citet{gaug} proposed to manipulate edges with a separate edge prediction module in training and inference stages. 
Another approach, GraphMix \cite{graphmix}, applies manifold Mixup \cite{manifold_mixup} to graph node classification which jointly trains a Fully Connected Network (FCN) and a GNN with shared parameters. 
 
Data augmentation for graph classification tasks, which is our main concern, is relatively less explored. \citet{graphcl} created new graphs with four general data augmentations: node dropping, edge perturbation, attribute masking, and subgraph extraction for contrastive learning. 
These baselines do succeed in enlarging the dataset size but take the risk of creating misleading data since the characteristics of each graph are not considered. In this sense, a filtering method for augmented data is proposed \cite{augforgraphclf} that compares the prediction value of augmented and validation data after augmenting with edge perturbation to decide if the augmented data are taken or not. \citet{wang2021mixup} also proposed Mixup-like augmentations for graph and node classification, respectively. For graph classification, they suggested to mixup final graph representation vectors after a readout layer, which can be viewed as a simple extension of manifold Mixup to graph data. 



\paragraph{Mixup approach}
The recently proposed Mixup~\cite{mixup} uses a pair of input samples to generate augmented data via convex combination. This approach demonstrates its effectiveness in that mixed data allows the model to learn a more general decision boundary and improve its generalization performance for unseen data. Moreover, Manifold Mixup~\cite{manifold_mixup} expands the Mixup to the latent feature space. Especially, a novel extension of Mixup called CutMix~\cite{cutmix} generates a virtual image by cutting an image patch from one image and pasting it to the other, and mixes their labels according to the area each occupies. This method effectively compensates the information loss issue in Cutout~\cite{cutout}, while taking advantage of the regional dropout enforcing the model to attend on the entire object. 
While Mixup and its variants lead the data augmentation field not only in CV but also in NLP \cite{nlpmixup} and ASR \cite{medennikov2018investigation}, the mixing data scheme is not yet transplanted to the domain of graphs. Since the nodes of the two different graphs do not match and the connectivity of the newly mixed graph must be determined, mixing two graphs is non-trivial and becomes a difficult challenge. 


\paragraph{Saliency}
Interpreting the model predictions has emerged as an important topic in the field of deep learning. Specifically, many researchers exploit the gradient of output with respect to input features to compute the saliency map of the image. In the domain of graphs, there are also attempts to identify the salient sub-structures or node features that primarily affect the model prediction \cite{gnnexplainer,neil2018interpretable}. 
The simple approach using the gradient of the loss with respect to the node features may somewhat be less accurate than other more advanced interpretability methods such as \cite{gnnexplainer}, but it does not require any network modifications as \cite{grad-cam} nor the training of additional networks \cite{zhou2016learning}. 

\section{Preliminary}
\paragraph{Graph classification} 

Let us define $G = (V, E)$ to be an undirected graph where $V$ is the set of nodes, $E$ is the set of edges or paired nodes. $X \in \mathbb{R}^{|V| \times d}$ is the feature matrix whose $i$-th entry is a $d$-dimensional node feature vector for node $i$. $\mathcal{N}(v)$ is the set of nodes $\{u\in V| \{u, v\} \in E\}$ that are connected to $v$ by edges. For the graph $C$-class classification task, each data point $(G, y)$ consists of a graph $G$ and its corresponding label $y \in \{1,\ldots, C\}$.  The goal is to train a GNN model which takes the graph $G$ and predicts the probability for each class.
\paragraph{Graph neural networks for graph classification} 

In this paper, we consider different members of message-passing GNN framework with the following four differentiable components: 1) message function $m_l$,  2) permutation invariant message aggregation function $\phi_l$ such as element-wise mean, sum or max, 3) node update function $h_l$, and 4) readout function $\gamma$ that permutation-invariantly combines node embeddings together, at the end to obtain graph representation \cite{gcnn, gatt, ginn}. Let  $x_v^{(l)}$  be the latent vector of node $v$ at layer $l$. To simplify notations with the recursive definition of GNNs, we use $x_v^{(0)}$ to denote the input node feature. At each GNN layer, node features are updated as $ x_v^{(l+1)} = h_l( x_v^{(l)}, \phi_i (\{m_l(x_v^{(l)}, x_u^{(l)}, e_{v, u})| u \in \mathcal{N}(v)\})$ and after the last $L$-th layer, all node feature vectors are combined by the readout function: $x_G = \gamma({x_0^{(L)}, ... , x_{|V|}^{(L)}})$. By passing $x_G$ to a classifier $f$, we obtain the prediction $\hat{y} = f(x_G)$ for graph $G$ where $\hat{y}_c = P(y=c|G)$. As a representative example, a layer of Graph Convolutional Network (GCN) is defined as $x_v^{(l+1)} = \Theta \sum_{u\in \mathcal{N}(v) \cup \{v\}} \frac{e_{v, u}}{\sqrt{\hat{d}_u \hat{d}_v}} x_u^{(l)} $ where  $\hat{d}_v =  1 + \sum_{u \in \mathcal{N}(v)} {e_{v, u}}$ with the edge weight $e_{v, u}$.
\paragraph{Mixup and its variants}

Mixup is an interpolation-based regularization technique that generates virtual data by combining pairs of examples and their labels for classification problems. Let $x \in \mathcal{X}$ be an input vector and $y \in \mathcal{Y}$ be an one-hot encoded $C$-class label from data distribution $\mathcal{D}$. Mixup based augmentation methods optimize the following loss $\mathcal{L}$ given a mixing ratio distribution $q$:
\begin{align}
\minimize_{\theta} \mathbb{E}_{(x_i, y_i), (x_j, y_j)\sim \mathcal{D}, \lambda \sim q}[\mathcal{L}(g(x_i, x_j; \lambda), \ell(y_i, y_j ; \lambda))] \nonumber
\end{align}
where $g$ and $\ell$ are the data and label Mixup functions, respectively. In case of the vanilla Mixup \cite{mixup}, these functions are defined as $g(x_i, x_j;\lambda) = (1-\lambda) x_i + \lambda x_j$  and $\ell(y_i, y_j) = (1-\lambda) y_i + \lambda y_j$ with the mixing parameter $\lambda$ sampled from beta distribution $\texttt{Beta}(\alpha, \alpha)$. 

Another example, CutMix \cite{cutmix}, defines mixing function as $g(x_i, x_j;B) = (1-\mathbbm{1}_B)\odot x_i + \mathbbm{1}_B\odot x_j$ where $\odot$ denotes the element-wise Hadamard product for a binary mask $\mathbbm{1}_B$ with $B={[a,a + w] \times [b, b+h]}$ and $\lambda = \frac{wh}{WH}$ where $W$ and $H$ are the width and height of images. 

\section{Graph Transplant}
\begin{figure*}[!tbp]
  \centering
  \includegraphics[width= 0.725\linewidth]{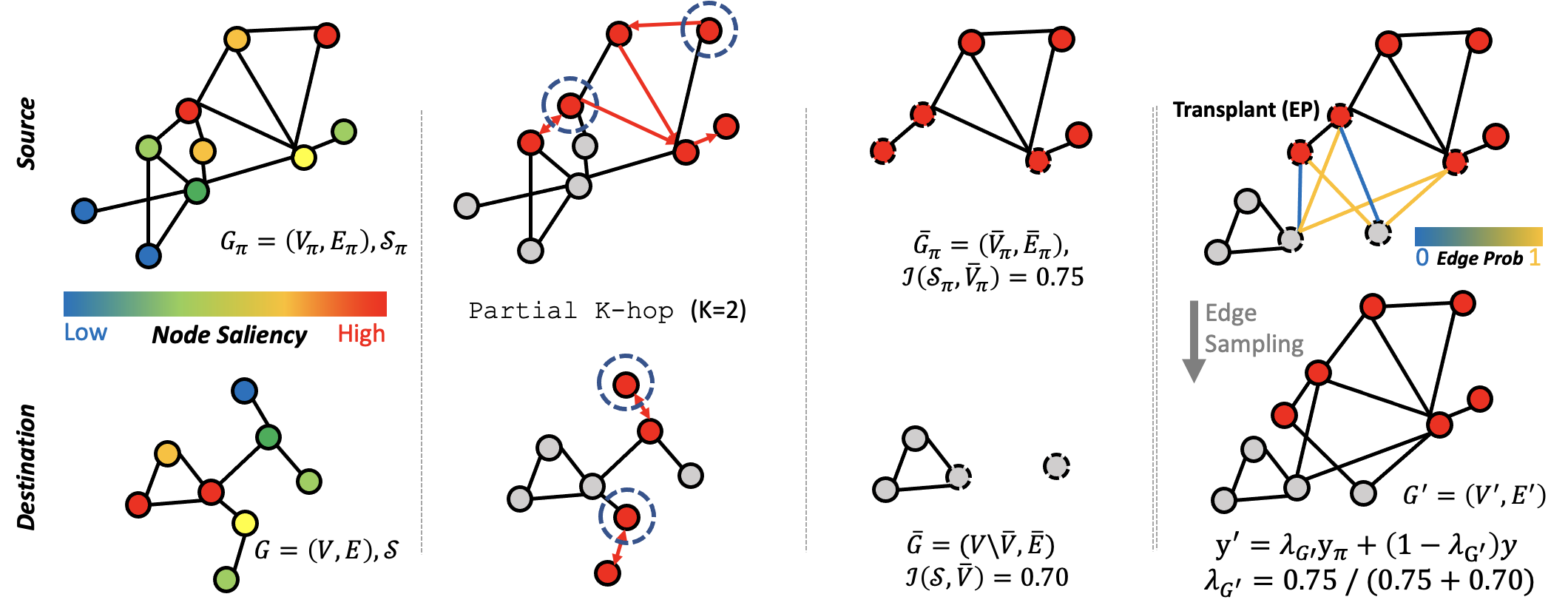}
  \caption{\small Overview of \emph{Graph Transplant}. First, we compute the node saliency vectors $\mathcal{S}_{\pi}, \mathcal{S}$ of source/destination graphs $G_\pi, G$. Then, we extract partial $K$-hop subgraphs from each graph anchored by salient nodes $\hat V_\pi$ and random nodes $\hat V$, respectively. Mixed label $y'$ is adaptively determined based on saliency  $\mathcal{S}_{\pi}, \mathcal{S}$. Finally, we transplant the source-side subgraph $\bar G_\pi$ to the remaining destination subgraph. 
  }
  \label{fig:concept}
  \vspace{-0.05in}
\end{figure*}


Now we introduce our effective augmentation strategy for graph classification, \emph{Graph Transplant}, which generates semantically meaningful mixed graphs considering the context of graph instances. First, we define the node saliency for GNNs that encodes the importance of each node in classifying the graph property (Section \ref{subsec:nodesaliency}). 
Armed with this notion of node saliency, we devise two main components of \emph{Graph Transplant}: graph mixing that preserves the local graph structure (Section \ref{subsec:graphmixing}) and the label mixing that adaptively assigns the supervisions based on  saliency of each subgraph to appropriately describe the mixed graphs (Section \ref{subsec:labelmixing}). The details are described below. 

\subsection{Node Saliency for GNNs} \label{subsec:nodesaliency}
\emph{Graph Transplant} does not rely on a specific method of calculating the importance of a subgraph and can be combined with a variety of approaches.
However, unlike the general purpose of explainability, which finds important regions given a trained model, our goal is to dynamically find and mix up subgraphs while GNN is being trained. Hence, in this paper, we preferentially consider an approach that can simply and quickly find the importance of a node using node features and gradients. 
Among them, in particular, motivated by the works only using gradient values for image classification \cite{simonyan,saliency_activation,puzzlemix}, we devise the following definition of the node saliency.

Given the latent node feature matrix $X^{(l)}$ obtained from the GNN $l$-th layer, we first compute the gradient of classification loss $\mathcal{L}(G, y=c) = -\log \hat{y}_c$ with respect to $X^{(l)}$, 
$\frac{\partial \mathcal{L}}{\partial X^{(l)}} \in \mathbb{R}^{|V| \times d}$. 
It is worth noting that the gradient can be obtained almost without cost as it is a byproduct of training step on the original graphs.
Then, the saliency $s_v$ of node $v$ is derived as the $\ell_2$ norm of the gradient corresponding to node $v$:
$s_v = \left|\left|\left[\frac{\partial \mathcal{L}}{\partial X^{(l)}}\right]_{v,:} \right|\right|_{2}.$
The choice of computing gradients at the $l$-th layer is to capture 
the high-level spatial information in $l$-hop neighborhood for each node 
\cite{bengio2013representation, grad-cam}. The final node saliency vector $\mathcal{S}_{G} = [s_1, ... , s_{|V|}]^{T}$ is therefore obtained without modifying the network architecture. 

\subsection{Graph Mixing} \label{subsec:graphmixing}

Graph mixing consists of the following two steps. 

\paragraph{Salient subgraph selection} 


For every iteration, we consider the batch of $N$ graphs $(G_1,\ldots, G_N)$. We pair two graphs $G_{\pi_i}$ and $G_i$ where $\pi$ is a shuffled index of $\texttt{range}(N) = [1, \ldots, N]$. From now on, we drop the index $i$ for brevity, and $G_{\pi}$ and $G$ will be referred to as a source and a destination graph, respectively. 


To construct the subgraph $\bar{G}_\pi = (\bar{V}_\pi, \bar{E}_\pi)$ from the source $G_{\pi}$, we first choose salient $R\%$ of nodes $\hat{V}_\pi$, to locate the important part containing the class semantics. The node saliency represents the importance of the $l$-hop subgraph by its definition. To avoid repeatedly selecting the same nodes for the whole training phase, we stochastically sample $\hat{V}_\pi$, from the top-$2R\%$ salient nodes 
uniformly at random. %
Given the set of selected anchors $\hat{V}_\pi$, we extract the \emph{partial} $K$-hop subgraphs $\bar{G}_\pi$ by randomly taking $p\%$ of adjacent neighbors for each moving step. $p$ is sampled from $\texttt{Beta}(\alpha, \alpha)* 100$ for each training iteration (See Algorithm~\ref{alg:paritial subgraph}). The main reason for choosing the partial subgraph is for the case of when the graph is dense, the $K$-hop subgraph may coincide with the whole graph $G$. For each iteration, $K$ is stochastically sampled from a discrete space $\mathcal{K}$.

On the destination side, the subgraph is removed from the original graph to `transplant' the source-side subgraph $\bar{G}_\pi$. The nodes $\bar{V}$ obtained by Algorithm~\ref{alg:paritial subgraph} are removed from destination graph $G$. In contrast to the source-side, anchor nodes $\hat{V}$ are \emph{randomly} selected at $R\%$ in the destination graph $G$ for the diversity of augmented graphs. The place where these nodes fall off is where the subgraph from the source, $\bar{G}_\pi$, will be transplanted. The remaining destination subgraph induced by $V \setminus \bar{V}$ is denoted by $\bar{G} = (V\setminus\bar{V}, \bar{E})$.
\begin{algorithm}[tbh]
\scriptsize
\caption{\footnotesize $\texttt{Partial K-hop}$}\label{alg:paritial subgraph}
\textbf{Input}: number of hops $K$, original graph $G$, subnodes $\hat{V} \subset V$, sampling ratio $p$ \\
\textbf{Output}: $\bar{V}$ \\
\textbf{Initialize}: $\bar{V} \leftarrow \hat{V}, H \leftarrow \hat{V}$, 
\begin{algorithmic}[1]
\FOR{$k=1$ to $K$}
    \STATE $H \leftarrow$ randomly chosen $p~\%$ of nodes in $\mathcal{N}(H)$ 
    \STATE $\bar{V} \leftarrow \bar{V} \cup H $ 
\ENDFOR
\end{algorithmic}
\end{algorithm}
\vspace{-0.15 in}

\begin{algorithm}[tbh]
\scriptsize
\caption{\footnotesize $\texttt{Graph Transplant}_{EP}$}\label{alg:graph grafting}
\textbf{Input}: graph pair $(G_\pi, G)$, node saliency $\mathcal{S}_{G_\pi}$, number of hops $K$, central nodes sampling ratio $R$, edge prediction module $\psi$ \\
\textbf{Output}: mixed graph $G' = (V', E')$ 

\begin{algorithmic}[1]

\STATE $\hat{V}_\pi \leftarrow$ $R\%$ salient nodes from $G_\pi$ given $\mathcal{S}_{G_\pi}$ \COMMENTmy{Source}
\STATE $\bar{V}_\pi \leftarrow$ $\texttt{Partial K-hop}$ $(K$, $G_\pi$, $\hat{V}_\pi)$
\STATE $\bar{G}_\pi\leftarrow$ Induced Subgraph $(G_\pi$, $\bar{V}_\pi)$
\STATE $\hat{V} \leftarrow$ $R\%$ random nodes from $G$ \COMMENTmy{Destination}
\STATE $\bar{V} \leftarrow$ $\texttt{Partial K-hop}$ $(K$, $G$, $\hat{V})$
\STATE $\bar{G} \leftarrow$ Induced Subgraph $(G$, $V\setminus\bar{V})$
\\  \LineComment{Merge}
\STATE $V' \leftarrow \bar{V}_\pi \cup (V \setminus \bar{V})$ 
\STATE $E' \leftarrow \bar{E}_\pi \cup \bar{E}$
\\  \LineComment{Edge prediction}
\FOR{$u \in \bar{V}_\pi$, $v \in V\setminus\bar{V}$} 
\IF{$\texttt{deg}_{G_\pi}(u) > \texttt{deg}_{\bar{G}_\pi}(u), \texttt{deg}_{G}(v) > \texttt{deg}_{\bar{G}}(v)$}
 \STATE $\bar{e}_{u, v} \sim \texttt{Bernoulli}\big(\frac{1}{2}\psi(u, v)+\frac{1}{2}\psi(v, u)\big)$
\ELSE
 \STATE $\bar{e}_{u, v} \leftarrow 0$
 \ENDIF
\ENDFOR
\STATE $E' \leftarrow E' \cup \{\{u, v\}|u\in \bar{V}_\pi, v\in V \setminus \bar{V}, \bar{e}_{u, v} = 1 \}$
\end{algorithmic}

\end{algorithm}

\paragraph{Transplant} \label{par:grafting}
After extracting the subgraphs $\bar{G}_\pi$ and $\bar{G}$, we transplant $\bar{G}_\pi$ into $\bar{G}$ by adding new edges between $\bar{V}_\pi $ and $V\setminus\bar{V}$. Formally, we construct the combined graph $G'=(V', E')$ with $V' = {\bar{V}}_{\pi} \cup (V \setminus {\bar{V}})$ and need to determine the connectivity between ${\bar{V}}_{\pi}$ and $(V \setminus {\bar{V}})$. We only consider the nodes losing edges by subgraph extraction to preserve the local information and the inner-structure intact.
Toward this, let us define $\texttt{deg}_G(v)$ as the degree of node $v$ in graph $G$. Then, 
the set of candidate nodes of the source and the destination graphs where the factitious edges can be attached is $U_\pi = \{v\in {\bar{V}}_{\pi} | \texttt{deg}_{G_\pi}(v) > \texttt{deg}_{\bar{G}_{\pi}}(v)\}$, $U = \{v\in (V \setminus {\bar{V}}) | \texttt{deg}_G(v) > \texttt{deg}_{\bar{G}}(v)\}$ respectively, and the total amount of change in degree is $D_\pi = \sum_{v \in U_\pi} (\texttt{deg}_{G_\pi} (v) - \texttt{deg}_{\bar{G}_\pi}(v))$, $D = \sum_{v \in U} (\texttt{deg}_G (v) - \texttt{deg}_{\bar{G}}(v))$. 

To decide the connectivity between $v_\pi \in U_\pi$ and $v\in U$, we propose the following two different approaches. The first one is to sample $\lfloor (D + D_\pi)/2 \rfloor$ node pairs $(u, v) \in U_\pi \times U$ uniformly at random. This straightforward and efficient method, which we call \emph{degree preservation} abbreviated as $DP$,
preserves the expected degree of the nodes in the original graphs. This makes the distribution of the generated graphs not too away from the original one while producing diverse samples. 
The other is to introduce a differentiable edge prediction module in the training phase to take into account the features of node pairs for connectivity. We name it \emph{edge prediction} or $EP$.
The edge prediction module $\psi$ takes a concatenated pair of node features $(r_{v_\pi}, r_v)$ and predicts the probability of edge existence $\psi(r_{v_\pi}, r_v) = P(\{v_\pi, v\} \in E')$. 
As a slight abuse of notation, we use $\psi(u, v)$ to denote the edge predictor consuming representations of node pair $(u, v)$. For undirected graphs, we consider the symmetric form and take the average $\hat{e}_{u, v} = \frac{1}{2} (\psi(u, v) + \psi(v, u))$. For inputs of $\psi$, we utilize latent node features $x_{v_\pi}^{(l)}$ and $x_v^{(l)}$ after $l$-th GNN layer to leverage the local information together. We sample the new edge weight $\bar{e}_{v_\pi, v}$ from $\texttt{Bernoulli}(\hat{e}_{v_\pi, v})$. To make the sampled edge weight $\bar{e}$ differentiable, we resort to the straight-through Gumbel-Softmax estimator \cite{gumbel-softmax} in the training phase. Finally, a new edge set $E'$ is set to $\bar{E}_\pi \cup \bar{E} \cup \{\{v_\pi, v\}| v_\pi \in \bar{V}_\pi, v \in V, \bar{e}_{v_\pi, v} = 1 \}$. Edge prediction module $\psi$ is trained by both supervised learning with the original graphs and in an end-to-end manner with virtual graphs (see Lines 4-5, 13 of Algorithm \ref{alg:graph grafting training}). In Supplementary, we provide the accuracy learning curve of the edge prediction module for node pairs of original graphs.

\begin{algorithm}[tbh]
\scriptsize
\caption{\footnotesize Train with $\text{\texttt{Graph Transplant}}_{EP}$ }\label{alg:graph grafting training}
\textbf{Input}: data distribution $\mathcal{D} = \{(G, y)\}$, GNN model $\theta$, central nodes sampling ratio $R$, edge prediction module $\psi$, discrete $K$-hop space $\mathcal{K}$, learning rate $\eta$
\begin{algorithmic}[1]
\WHILE{not satisfying early stopping criteria}
\STATE Sample batch $\mathcal{B} = \{(G_i, y_i)\} \sim \mathcal{D}$
\STATE Backpropagate to obtain $\nabla_\theta\mathcal{L} = \nabla_\theta \frac{1}{|\mathcal{B}|} \sum_i \mathcal{L} (G_i, y_i)$ and node saliency $\{\mathcal{S}_{G_i}\}_i$ given $\theta$ 
\\ \LineComment{Supervised update of the edge prediction module}
\STATE Sample pairs of node representation and their connection \\ $\mathcal{E} = \{((x_u^{(l)}, x_v^{(l)}), \mathbbm{1}_{\{u, v\} \in \cup_i E_i})\}$ 
\STATE Update the edge predictor $\psi$ with $\mathcal{E}$ 
\\ \LineComment{Do \emph{Graph Transplant}}
\STATE Pair graphs by shuffling batch $\{(G_{\pi_i}, G_i)\}_i$
\STATE $K \leftarrow \texttt{random.choice}(\mathcal{K})$
\FOR{$i = 1$ to $|\mathcal{B}|$ }
\STATE  $G'_i \leftarrow \scriptsize{\texttt{Graph Transplant}}_{EP}\big((G_{\pi_i}, G_i)\big.$, $\mathcal{S}_{G_{\pi_i}}$,$K$,$R$,$\big.\psi\big)$ 
\STATE  $y'_i \leftarrow \ell(y_{\pi_i}, y_i)$ 
\ENDFOR
\\ \LineComment{Update}
\STATE $\theta \leftarrow \theta - \eta(\nabla_\theta \mathcal{L} + \nabla_\theta \frac{1}{|\mathcal{B}|} \sum_i \mathcal{L}_{Aug}(G'_i, y'_i))$
\STATE $\psi \leftarrow \psi - \eta\nabla_\psi \frac{1}{|\mathcal{B}|} \sum_i \mathcal{L}_{Aug}(G'_i, y'_i)$
\ENDWHILE
\end{algorithmic}
\end{algorithm}

\subsection{Adaptive Label Mixing} \label{subsec:labelmixing}
In the previous section, our method generates virtual graph data that captures the graph semantics while preserving the original local information. However, assigning the appropriate label to the generated graph is necessary to prevent the network from being misled by inappropriate supervision from noisy data. For example, suppose that the core nodes that mainly determine the property of destination graph $G$ are condensed in a small region. If the nodes in this small region are eliminated via our subgraph selection method described above, the original property of $G$ is mostly lost, and hence simply assigning the label in proportion to the number of nodes will not explain the generated graph properly. To tackle this issue, we propose a simple but effective label mixing strategy for \emph{Graph Transplant}.

We define the importance of source subgraph $\bar{G}_\pi$ 
(and the remaining part of destination graph $\bar{G}$) as the total saliency of the nodes constituting the subgraph in the original full graph. This adaptive importance function $\mathcal{I}({\mathcal{S}, V})$ can be formulated as 
$\mathcal{I}({\mathcal{S}, V}) = \frac{\sum_{v \in V}{s_v}}{\|\mathcal{S}\|_{1}}$.
The label weight $\lambda_{G'}$ for generated graph $G'$ is derived according to the relative importance of the parts being joined from the source $\mathcal{I}({\mathcal{S}_{G_\pi}, \bar{V}_\pi})$ and the destination $\mathcal{I}({\mathcal{S}_{G}, V \setminus \bar{V}})$, so that the label for the mixed instance $\ell(y_\pi, y)$ is defined as $\lambda_{G'} y_\pi + (1-\lambda_{G'})y$ where
$\lambda_{G'} = \frac{\mathcal{I}({\mathcal{S}_{G_\pi}, \bar{V}_\pi})}{\mathcal{I}({\mathcal{S}_{G_\pi}, \bar{V}_\pi})+\mathcal{I}({\mathcal{S}_{G}, V \setminus \bar{V}})}$.

Our final objective is 
\begin{align}\label{eq:objective}
\minimize_{\theta, \psi} \mathbb{E}_{(G_\pi, y_\pi), (G, y)\sim \mathcal{D}}[\mathcal{L} (G, y) + \mathcal{L}_{Aug}(G', y')]
\end{align}
where $\mathcal{L}_{Aug}(G', y')$ is defined as $ \lambda_{G'}\mathcal{L}(G', y_\pi) + (1-\lambda_{G'})\mathcal{L}(G', y)$. The whole learning process with \emph{Graph Transplant} is described in Algorithm~\ref{alg:graph grafting training}. 

    

\begin{table*}[t]
\caption{\small \BF{(Left)} To compare `$\emph{Graph Transplant}_{DP/EP}$' with the other baselines, we report the averaged accuracy and the standard deviation of 3 repeats of 5-fold cross validation on five graph classification benchmark datasets (\textit{Small-scale}). 
Results of GCN are deferred to Supplementary. Note that MaskN can not be applied to COLLAB which does not have node features.  \BF{(Right)} Our method also enhances the classification performance on the larger-scale datasets (\textit{Medium-scale}). We report the mean and standard deviation of AUROC for 5-fold. } 
\vspace{-0.15in}
\begin{center}
\setlength{\tabcolsep}{6.0pt} 
\begin{scriptsize}
\begin{adjustbox}{width=1.\linewidth}
\begin{tabular}{@{\extracolsep{8pt}}rlccccc||ccc@{}}
\toprule
& \multirow{2}{*}{\textbf{Method}} & \multicolumn{5}{c}{\textbf{Dataset (\textit{Small-scale})}} & \multicolumn{3}{c}{\textbf{Dataset (\textit{Medium-scale})}} \\
\cline{3-7}\cline{8-10}
&  & COLLAB & NCI1 & ENZYMES & Mutagenicity & COIL-DEL & NCI-H23 & MOLT-4 & P388 \\
\cline{2-10}
\rule{0pt}{2.5ex}  

\multirow{9}{*}{\rotatebox{90}{GCS}} & Vanilla & 0.814 \tiny{$\pm 0.010$} & 0.784 \tiny{$\pm 0.025$} 
& 0.639 \tiny{$\pm 0.040$} & 0.819 \tiny{$\pm 0.009$}& 0.688 \tiny{$\pm 0.019$} & 0.876 \tiny{$\pm 0.009$} & 0.833 \tiny{$\pm 0.011$} & 0.905 \tiny{$\pm 0.009$} \\
                    \cdashline{2-10}
                    \rule{0pt}{2ex}
                     & Manifold Mixup & 0.814 \tiny{$\pm 0.010$}  &  0.792 \tiny{$\pm 0.018$} 
                     & 0.664 \tiny{$\pm 0.053$}  &  0.822 \tiny{$\pm 0.009$}& 0.756 \tiny{$\pm 0.029$} & 0.884 \tiny{$\pm 0.013$}  &  \BF{0.842} \tiny{$\pm 0.011$} & 0.902 \tiny{$\pm 0.021$} \\
                     & M-Evolve & 0.817 \tiny{$\pm 0.011$} & 0.775 \tiny{$\pm 0.023$} 
                      & 0.666 \tiny{$\pm 0.059$}  & 0.821 \tiny{$\pm 0.009$} &  0.703 \tiny{$\pm 0.013$}  & 0.875 \tiny{$\pm 0.009$} & 0.829 \tiny{$\pm0.011$} & 0.904 \tiny{$\pm 0.011$} \\
                     & DropN & 0.816 \tiny{$\pm 0.008$}  & 0.789 \tiny{$\pm 0.031$} 
                     & 0.686 \tiny{$\pm 0.056$} & 0.820 \tiny{$\pm 0.018$} & 0.860 \tiny{$\pm 0.021$} & 0.878 \tiny{$\pm 0.012$}  & 0.834 \tiny{$\pm 0.011$} & 0.911 \tiny{$\pm 0.013$}  \\
                     & PermE & 0.816 \tiny{$\pm 0.006$} & 0.785 \tiny{$\pm 0.023$} 
                     & 0.675 \tiny{$\pm 0.052$} & 0.820 \tiny{$\pm 0.015$} & 0.796 \tiny{$\pm 0.016$} & 0.884 \tiny{$\pm 0.004$} & 0.840 \tiny{$\pm 0.010$} & 0.910 \tiny{$\pm 0.014$}\\
                     & MaskN &-& 0.790 \tiny{$\pm 0.018$}  
                     &  0.678 \tiny{$\pm 0.046$} & 0.819 \tiny{$\pm 0.008$} & 0.791 \tiny{$\pm 0.021$}  & 0.877 \tiny{$\pm 0.006$}& 0.838 \tiny{$\pm 0.007$} & 0.907 \tiny{$\pm 0.010$}  \\
                     & SubG & 0.812 \tiny{$\pm 0.010$} & 0.784 \tiny{$\pm 0.021$} 
                     & 0.675 \tiny{$\pm 0.057$} &  0.821 \tiny{$\pm 0.011$} & 0.819 \tiny{$\pm 0.025$} & 0.877 \tiny{$\pm 0.012$} & 0.836 \tiny{$\pm 0.011$} & 0.911 \tiny{$\pm 0.014$} \\
                    
                     \cline{2-10}
                     \rule{0pt}{2ex}
                      & \textbf{$\text{Graph Transplant}_{DP}$} & 0.820 \tiny{$\pm 0.011$} & \BF{0.815} \tiny{$\pm 0.011$} 
                     & 0.694 \tiny{$\pm 0.065$} &  0.821 \tiny{$\pm 0.010$} & 0.858 \tiny{$\pm 0.022$} & 0.880 \tiny{$\pm 0.016$} & \BF{0.843} \tiny{$\pm 0.003$} & \BF{0.921} \tiny{$\pm 0.010$}\\
                     
                     & \textbf{$\text{Graph Transplant}_{EP}$} & \BF{0.828} \tiny{$\pm 0.007$} & 0.807 \tiny{$\pm 0.013$} 
                     & \BF{0.707} \tiny{$\pm 0.053$} &  \textbf{0.829} \tiny{$\pm 0.010$} & \BF{0.865} \tiny{$\pm 0.016$}  & \BF{0.890} \tiny{$\pm 0.018$} & \BF{0.843} \tiny{$\pm 0.009$} 
                     & 0.919 \tiny{$\pm 0.007$} \\
                     
\cline{2-10}
\noalign{\vskip\doublerulesep
         \vskip-\arrayrulewidth} \cline{2-10}
\rule{0pt}{2.5ex}  
\multirow{9}{*}{\rotatebox{90}{GIN}} & Vanilla & 0.805 \tiny{$\pm 0.007$} & 0.793 \tiny{$\pm 0.013$} 
& 0.596 \tiny{$\pm 0.030$} & 0.815 \tiny{$\pm 0.010$}& 0.620 \tiny{$\pm 0.013$} & 0.879 \tiny{$\pm 0.006$} & 0.834 \tiny{$\pm 0.015$} & 0.901 \tiny{$\pm 0.012$} \\
                     \cdashline{2-10}
                     \rule{0pt}{2ex}
                    & Manifold Mixup & 0.809 \tiny{$\pm0.008$}  &  0.797 \tiny{$\pm 0.013 $} 
                     &  0.618 \tiny{$\pm 0.042$}  & 0.822 \tiny{$\pm 0.011$} &  0.654 \tiny{$\pm 0.020 $} & 0.873 \tiny{$\pm0.015$}  &  0.824 \tiny{$\pm 0.009 $} & 0.902 \tiny{$\pm 0.006$} \\
                     & M-Evolve & 0.809 \tiny{$\pm 0.007 $} & 0.785 \tiny{$\pm 0.015$} 
                      & 0.608 \tiny{$\pm 0.031$}  & 0.813 \tiny{$\pm 0.011$} &  0.604 \tiny{$\pm 0.017$}  & 0.877 \tiny{$\pm 0.005$} &  0.832 \tiny{$\pm 0.016$} & 0.895 \tiny{$\pm 0.013$}\\
                     & DropN & 0.807 \tiny{$\pm 0.014$} & 0.789 \tiny{$\pm 0.022$}  
                     &  0.619 \tiny{$\pm 0.045$} &  0.822 \tiny{$\pm 0.010$} & 0.665 \tiny{$\pm 0.030$} & 0.850 \tiny{$\pm 0.030$} & 0.821 \tiny{$\pm 0.009$}  &  0.896 \tiny{$\pm 0.012$} \\
                     & PermE & 0.809 \tiny{$\pm 0.008$} & 0.790 \tiny{$\pm 0.018$}  
                     & 0.641 \tiny{$\pm 0.038$} & 0.823 \tiny{$\pm 0.008$} &  0.689 \tiny{$\pm 0.023$} & 0.867 \tiny{$\pm 0.012$} & 0.836 \tiny{$\pm 0.011$} & 0.910 \tiny{$\pm 0.018$}\\
                     & MaskN &- &  0.794 \tiny{$\pm 0.017$} 
                     &  0.624 \tiny{$\pm 0.028$}& 0.820 \tiny{$\pm 0.008$}  &  0.632 \tiny{$\pm 0.017$} & 0.877 \tiny{$\pm 0.007$} &  0.831 \tiny{$\pm 0.014$} &  0.898 \tiny{$\pm 0.017$} \\
                     & SubG & 0.805 \tiny{$\pm 0.010$} & 0.789 \tiny{$\pm 0.021 $} 
                     & 0.631 \tiny{$\pm 0.030$} &0.820 \tiny{$\pm 0.008$} & 0.667 \tiny{$\pm 0.022$}  & 0.878 \tiny{$\pm 0.005$} & 0.834 \tiny{$\pm 0.014 $} & 0.900 \tiny{$\pm 0.010$}\\
                  
                     \cline{2-10}
                     \rule{0pt}{2ex} 
                     & \textbf{$\text{Graph Transplant}_{DP}$} & \BF{0.814} \tiny{$\pm 0.012$} & \BF{0.815} \tiny{$\pm 0.011$} 
                     & \BF{0.653} \tiny{$\pm 0.038$} & 0.822 \tiny{$\pm 0.012$} & \BF{0.700} \tiny{$\pm 0.025$} & \BF{0.898} \tiny{$\pm 0.018$} & \BF{0.855} \tiny{$\pm 0.013$} & 0.914 \tiny{$\pm 0.009$}  \\
                     
                     & \textbf{$\text{Graph Transplant}_{EP}$} & \BF{0.815} \tiny{$\pm 0.009$} & 0.811 \tiny{$\pm 0.011$} 
                     & 0.649 \tiny{$\pm 0.039$} & \BF{0.828} \tiny{$\pm 0.009$} & 0.682 \tiny{$\pm 0.025$} & 0.890 \tiny{$\pm 0.012$} & 0.846 \tiny{$\pm 0.011$} 
                     & \BF{0.916} \tiny{$\pm 0.010$}  \\

\cline{2-10}
\noalign{\vskip\doublerulesep
         \vskip-\arrayrulewidth} \cline{2-10}
\rule{0pt}{2.5ex}  

\multirow{9}{*}{\rotatebox{90}{GAT}} & Vanilla & 0.804 \tiny{$\pm 0.013$} & 0.750 \tiny{$\pm 0.017$} 
& 0.415 \tiny{$\pm 0.190$} & 0.806 \tiny{$\pm 0.009$}& 0.680 \tiny{$\pm 0.020$} \\
                     \cdashline{2-7}
                     \rule{0pt}{2ex}
                      & Manifold Mixup & 0.805 \tiny{$\pm 0.010$} & 0.778 \tiny{$\pm 0.014$}  
                     &  0.507 \tiny{$\pm 0.225$} & 0.811 \tiny{$\pm 0.011$}&  0.741 \tiny{$\pm 0.023 $}  \\
                     & M-Evolve & 0.809 \tiny{$\pm 0.008$} & 0.728 \tiny{$\pm 0.023$} 
                      & 0.441 \tiny{$\pm 0.224$}  & 0.811 \tiny{$\pm 0.009$} & 0.729  \tiny{$\pm 0.027$}\\
                     & DropN & 0.802 \tiny{$\pm0.012$} & 0.752 \tiny{$\pm 0.038$} 
                     & 0.477 \tiny{$\pm 0.260$} & 0.810 \tiny{$\pm 0.013$} & \BF{0.811} \tiny{$\pm 0.018$}\\
                     & PermE & 0.804 \tiny{$\pm 0.009$} & 0.760 \tiny{$\pm 0.037$} 
                     & 0.532 \tiny{$\pm 0.196$} & 0.814 \tiny{$\pm 0.015$} & 0.791 \tiny{$\pm 0.019$}\\
                     & MaskN &- &0.756 \tiny{$\pm 0.028$}  
                     &  0.493 \tiny{$\pm 0.226$}& 0.808 \tiny{$\pm 0.012$} &  0.772 \tiny{$\pm 0.029$} \\
                     & SubG &0.792 \tiny{$\pm 0.008$} & 0.755 \tiny{$\pm 0.020$} 
                     & 0.506 \tiny{$\pm 0.220 $} &  0.815 \tiny{$\pm 0.012$} &0.731 \tiny{$\pm 0.022 $} \\
                    
                     \cline{2-7}
                     \rule{0pt}{2ex}  
                     & \textbf{$\text{Graph Transplant}_{DP}$} & \BF{0.813} \tiny{$\pm 0.010$} & \BF{0.793} \tiny{$\pm 0.018$} 
                     & 0.527 \tiny{$\pm 0.232$} & 0.815 \tiny{$\pm 0.009$} & 0.796 \tiny{$\pm 0.021$} \\
                     
                     & \textbf{$\text{Graph Transplant}_{EP}$} & \BF{0.813} \tiny{$\pm 0.009$} & 0.789 \tiny{$\pm 0.025$} 
                     & \textbf{0.551} \tiny{$\pm 0.220$} & \BF{0.817 \tiny{$\pm 0.011$}} & 0.795 \tiny{$\pm 0.022$} \\


\cline{2-7}
\noalign{\vskip\doublerulesep
         \vskip-\arrayrulewidth} \cline{1-7}

\end{tabular}
\end{adjustbox}
\vspace{-0.18 in}
\end{scriptsize}
\end{center}
\label{tb:maintable}
\end{table*}

\section{Experiment}
We perform a series of experiments on multiple domain datasets to verify our method \emph{Graph Transplant}. Since this is the first input-level Mixup-based work on graph-structured data themselves, we compare the performance against Manifold Mixup \cite{wang2021mixup} that mixes up the hidden representations after the pooling layer, and other stochastic graph augmentations (Section \ref{subsec:baselines}). 
Although theoretically \emph{Graph Transplant} can be unified to any GNN architecture without network modification, we conduct the experiments with four different GNN architectures
: GCN~\cite{gcnn}, GCS (GCN with learnable Skip connection), GAT~\cite{gatt}, GIN~\cite{ginn}. We also conduct ablation studies (Section \ref{subsec:ablation}) to explore the contribution of each component of our method - node saliency, saliency-based label mixing, and locality preservation. Finally, we conduct the experiments of model calibration and adversarial attack (Section~\ref{subsec:attack}, \ref{subsec:analysis}).

\subsection{Baselines} \label{subsec:baselines}
We now introduce the graph augmentation baselines to compare with \emph{Graph Transplant}. In addition to graph augmentation works, we also adopt general graph augmentations - DropN, PermE, MaskN, and SubG - from the recent graph contrastive learning work called GraphCL~\cite{graphcl} as our baselines. Since the proposed methods of this paper are catered for unsupervised learning, care was taken in crafting them into versions that enable data augmentation under supervised learning regime by assigning original labels to the augmented graphs. Note that for all baselines, we train the models with both original and augmented graphs for fair comparisons. We defer the details to Supplementary.

\paragraph{Manifold Mixup~\cite{wang2021mixup} }

Since the node feature matrix of each graph is compressed into one hidden vector after the readout layer $\gamma$, we can mix these graph hidden vectors and their labels. 

\vspace{-0.00in}

\paragraph{M-Evolve~\cite{augforgraphclf}} 
This method first perturbs the edges with a heuristically designed open-triad edge swapping method. Then, noisy-augmented graphs are filtered out based on the prediction of validation graphs. We tweak the framework a bit so that augmentation and filtration are conducted for every training step and filtration criterion is updated after each training epoch. 

\paragraph{Node dropping (DropN)}
Similar to Cutout \cite{cutout}, DropN randomly drops a certain portion of nodes. 
\vspace{-0.05in}
\paragraph{Edge perturbation (PermE)}
Manipulating the connectivity of the graph is one of the most common strategies for graph augmentation. 
PermE randomly adds and removes a certain ratio of edges. 
\vspace{-0.05in}
\paragraph{Attribute masking (MaskN)}
In line with Dropout \cite{srivastava2014dropout}, MaskN stochastically masks a certain ratio of node features. 
\vspace{-0.05in}
\paragraph{Subgraph (SubG)}
Using the random walk, SubG extracts the subgraph from the original graph. 

\subsection{Experimental Settings} \label{subsec:experiment_setting}

\paragraph{Datasets}
To demonstrate that \emph{Graph Transplant} brings consistent improvement across various domains of graphs and the number of graph instances 
, we conduct the experiment on 8 graph benchmark datasets: COLLAB~\cite{socialdataset} 
for social networks dataset, ENZYMES~\cite{enzymes}, obgb-ppa~\cite{hu2020ogb}
for bioinformatics dataset, COIL-DEL~\cite{coil-del} for computer vision dataset, and NCI1~\cite{nci1}, Mutagenicity~\cite{mutagen}, NCI-H23, MOLT-4, P388 ~\cite{large_mol} for molecules datasets. 


\paragraph{Evaluation protocol} 
We evaluate the model with 
5-fold cross validation. 
For datasets with less than 10k instances, the experiments are repeated three times so the total of 15 different train/validation/test stratified splits with the ratio of 3:1:1. 
We carefully design the unified training pipeline that can guarantee sufficient convergence. For small  and medium-scale datasets, we train the GNNs for 1000 epochs under the early stopping condition and terminate training when there is no further increase in validation accuracy for 1500 iterations. Similarly, the learning rate is decayed by 0.5 if there is no improvement in validation loss for 1000 iterations. Detailed evaluation protocols and implementation details of ours are provided in Supplementary. 




\begin{figure*}[!tbp]
  \centering
   \begin{minipage}[b]{0.27\textwidth}
    \includegraphics[width=\textwidth]{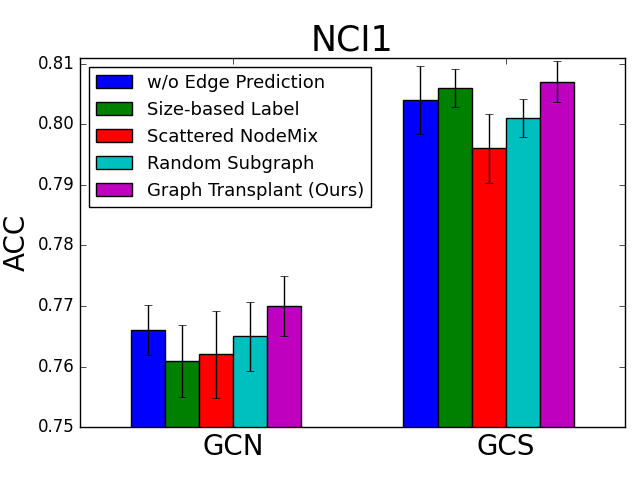}
  \end{minipage}
  \begin{minipage}[b]{0.27\textwidth}
    \includegraphics[width=\textwidth]{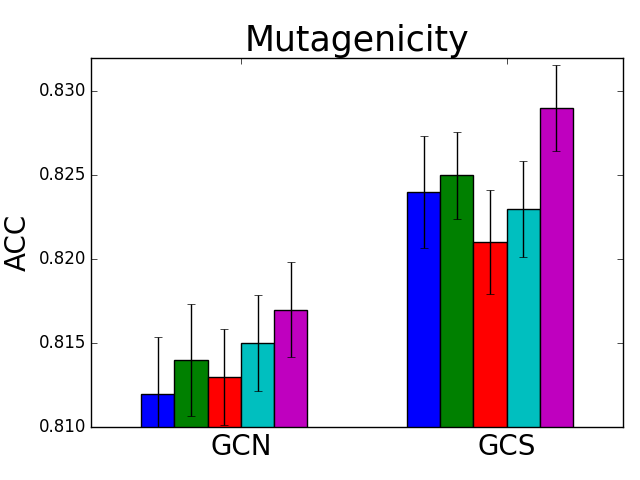}
  \end{minipage}
  \begin{minipage}[b]{0.265\textwidth}
    \includegraphics[width=\textwidth]{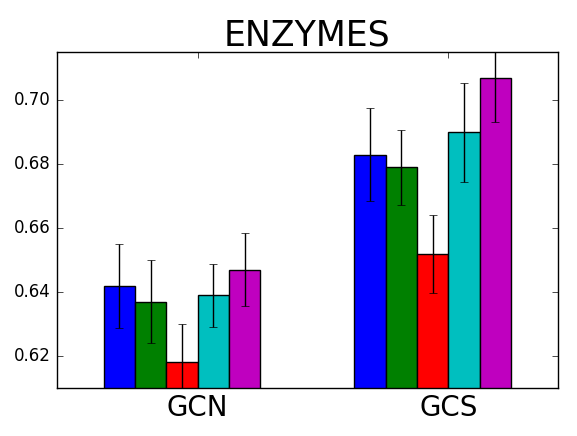}
  \end{minipage}
  \vspace{-0.15in}
  \caption{\small Ablation study. We draw a bar chart for each dataset, where bars are grouped by model architectures within a chart. The different ablation studies are distinguishable by colors. The rightmost bars which show the best performance are of our complete method.}
  \label{fig:ablation}
  \vspace{-0.18in}
\end{figure*}


\subsection{Main Results}
We report the averaged accuracy with standard deviation for the baselines and ours on the five datasets (Table~\ref{tb:maintable}, Left). Subscripts $DP $ and $EP$ denote transplant methods (See \BF{Transplant} in Section \ref{par:grafting}). \emph{Graph Transplant} achieves the best performances at the most cases. It is worth noting that, our method consistently exhibits superiority for datasets with different characteristics such as sparsity of connection, graph size, and the presence or absence of node features, while other baselines are jagged. Interestingly, our method with simple $DP$ generally beats the other baselines and is even superior to our method transplanting with $EP$ in some cases. We conjecture that it helps to allow the mixed graphs to be variety within the weak but reasonable constraint on the expected node degree. We obtain further gain in several cases by exploiting the edge predictor that utilizes the node features to determine the connectivity.

We also verify \emph{Graph Transplant} has advantages with larger datasets. We test on molecular datasets with 40k samples (Table~\ref{tb:maintable}, Right) with the two superior architectures  - GCS, GIN - and ogbg-ppa with 150k samples in bioinformatics (Table~\ref{tb:super_large}) with GCS. 
For class-imbalanced medium-scale datasets, we use AUROC for evaluation. 

\begin{table}[t]
\caption{\small We report the averaged accuracy with standard deviation for 3 runs of different data augmentation methods on \textit{ogbg-ppa}.}
\vspace{-0.1in}
\begin{center}
\setlength{\tabcolsep}{.5pt} 
\begin{footnotesize}
\begin{adjustbox}{width=0.668\linewidth}
\begin{tabular}{@{\extracolsep{8pt}}lc@{}}
\toprule
\textbf{Method} & \textbf{Open Graph Benchmark (OGB)} \\
\cline{2-2}
\textbf{(GCS)} & \textit{ogbg-ppa} \\
\noalign{\vskip\doublerulesep
         \vskip-\arrayrulewidth} \cline{1-2}
Vanilla & 0.638 \tiny{$\pm 0.007$} \\
                    \cdashline{1-2}
                     Manifold Mixup & 0.616 \tiny{$\pm 0.013$} \\
                     DropN & 0.660 \tiny{$\pm 0.005$} \\
                     PermE & 0.655 \tiny{$\pm 0.003$} \\
                     MaskN & 0.627 \tiny{$\pm 0.012$} \\
                     SubG & 0.661 \tiny{$\pm 0.005$} \\
                    
                     \cline{1-2}
                     
                     \textbf{$\text{Graph Transplant}_{EP}$} & \BF{0.680} \tiny{$\pm 0.002$} \\

\bottomrule
\end{tabular}
\end{adjustbox}
\end{footnotesize}

\end{center}
\label{tb:super_large}
\vspace{-0.2in}
\end{table}

\subsection{Ablation Study} \label{subsec:ablation}
To validate each component of \emph{Graph Transplant}, we conduct the following four ablation studies: \textbf{1)} Edge Prediction \textbf{2)} Label mixing strategies: Size-based vs Saliency-based \textbf{3)} Comparision of mixing units: $K$-hop subgraph vs scattered node\textbf{ 4)} Importance of saliency-based subgraph. The results are shown in Figure~\ref{fig:ablation}. Each color of the bars represents each variant of our method. The results for other datasets and detailed descriptions for each ablation study are deferred to Supplementary due to the space constraint. 

\subsection{Robustness to Adversarial Examples}~\label{subsec:attack}
Recently, mixup-variants prove their effectiveness in terms of model robustness~\cite{augmix, adv_mixup}. We test if the GNNs trained with \emph{GraphTransplant} can enhance the model robustness against adversarial attacks. We attack the GNNs using gradient-based white-box attack called \textit{GradArgMax}~\cite{gradargmax} designed for graph-structured data. In Table~\ref{tb:adversarial}, our method shows superior defense performance compared to the other baselines. Interestingly, albeit the adversarial attack is based on edge modification, ours exhibits better performance than PermE, an edge perturbation augmentation. Note that we get consistent results for other architectures. Detailed settings and additional results are described in Supplementary.
\begin{table}[h]
\vspace{-0.03in}
\caption{\small Comparison of robustness against adversarial attack.}
\vspace{-0.15in}
\setlength{\columnsep}{1pt}
\begin{center}
\begin{scriptsize}
\setlength{\tabcolsep}{0.8pt} 
\renewcommand{\arraystretch}{1} 
 \resizebox{0.8\linewidth}{!}{

\begin{tabular}{@{\extracolsep{3pt}}lccc@{}}
\toprule
\textbf{Method} & \multicolumn{3}{c}{\textbf{\textit{GradArgmax}} \scriptsize(White-Box Attack) }  \\ 
\cline{2-4}
\textbf{(GCS)} & ENZYMES & Mutagenicity & NCI1  \\
\cline{1-4}
                     Vanilla & 0.543 \tiny{$\pm 0.030$} & 0.579 \tiny{$\pm 0.013$}& 0.596 \tiny{$\pm 0.027$}\\
                     \cdashline{1-4}
                      Manifold Mixup & 0.575 \tiny{$\pm 0.041$}  & 0.596 \tiny{$\pm 0.019$} & 0.609 \tiny{$\pm 0.030$}\\
                      M-Evolve & 0.593 \tiny{$\pm 0.030$}  & 0.603 \tiny{$\pm 0.042$}  & 0.628 \tiny{$\pm 0.028$} \\
                     DropN & 0.617 \tiny{$\pm 0.030$} &  0.617 \tiny{$\pm 0.030$}& 0.631 \tiny{$\pm 0.028$} \\
                     PermE & 0.623 \tiny{$\pm 0.071$} & 0.645 \tiny{$\pm 0.031$}  & 0.602 \tiny{$\pm 0.042$}     \\
                     MaskN & 0.588\tiny{$\pm 0.050$}  & 0.514 \tiny{$\pm 0.024$} & 0.589 \tiny{$\pm 0.020$}       \\
                     SubG &  0.608 \tiny{$\pm 0.030$} & 0.582 \tiny{$\pm 0.024$} & 0.533 \tiny{$\pm 0.024$}       \\
                     \cline{1-4}
                     $\text{\textbf{Graph Transplant}}_{EP}$ & \BF{0.648} \tiny{$\pm 0.058$} & \BF{0.672} \tiny{$\pm 0.010$} & \BF{0.642} \tiny{$\pm 0.028$}   \\
\bottomrule
\end{tabular}
}
\end{scriptsize}
\end{center}
\label{tb:adversarial}
\vspace{-0.2in}
\end{table}

\subsection{Further Analysis}~\label{subsec:analysis}
\vspace{-0.18in}
\paragraph{Model Calibration} 
In image classification tasks, DNNs trained with mixup are known to significantly improves the model calibration. We evaluate if \emph{Graph Transplant} can alleviate the over/under-confident prediction of GNNs. In Table~\ref{tb:ece}, \emph{Graph Transplant} has the lowest Expected Calibration Error (ECE) in most cases. This implies that the GNNs trained with \emph{GraphTransplant} classifies the graphs with more accurate confidence, which is the better indicator of the actual likelihood of a correct prediction.

\begin{table}[h]
\caption{\small Comparison of the model calibration on GCS and GIN.}
\vspace{-0.15in}
\setlength{\columnsep}{1pt}
\begin{center}
\begin{scriptsize}
\setlength{\tabcolsep}{0.8pt} 
\renewcommand{\arraystretch}{1} 
\resizebox{0.75\linewidth}{!}{
\begin{tabular}{@{\extracolsep{3pt}}lccc@{}}

\toprule
\textbf{Method} & \multicolumn{3}{c}{\textbf{Expected Calibration Error}} \\ 
\cline{2-4}
\textbf{(GCS/GIN)} & ENZYMES & Mutagenicity & COIL-DEL \\
\cline{1-4}
 
                     Vanilla & 0.299/0.361 &  0.068/0.093 & 0.088/0.021   \\
                     \cdashline{1-4}
                     Manifold Mixup & 0.144/0.139  & 0.063/0.093  &  0.313/0.241   \\
                     M-Evolve & 0.285/0.319 & 0.056/0.069 & 0.065/0.099 \\
                     DropN & 0.252/0.327  & 0.051/0.080 & 0.051/0.327  \\
                     PermE & 0.242/0.301  & 0.052/0.069 & 0.062/0.327  \\
                     MaskN & 0.254/0.320  & 0.063/0.081 & \BF{0.046}/0.138  \\
                     SubG & 0.269/0.329   & 0.069/0.095 & 0.050/0.122  \\
                     
                     \cline{1-4}
                     $\text{\textbf{Graph Transplant}}_{EP}$ & \BF{0.046/0.136} & \BF{0.041/ 0.061} & 0.051/\BF{0.061}  \\
\bottomrule
\end{tabular}
}
\end{scriptsize}
\end{center}
\label{tb:ece}
\vspace{-0.2in}
\end{table}

\vspace{-0.02in}
\paragraph{Qualitative analysis}
We analyze the node saliency of the molecule graphs in Mutagenicity to show that our method improves generalizability and leads the model to attend where it really matters to correctly classify molecules. The task is to decide whether the molecule causes DNA mutation or not, which is important to capture the interaction of functional groups in molecules \cite{luch2005nature}.  

\begin{figure}[h]
  \centering
  \includegraphics[width=0.995\linewidth]{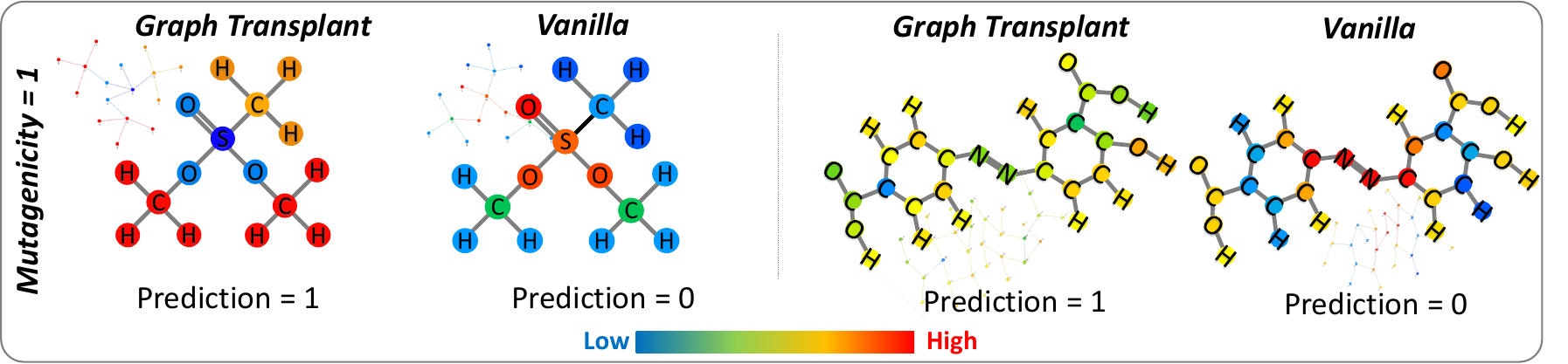}
  \vspace{-0.1in}
  \caption{\small Comparison of saliency from the trained models with \emph{Graph Transplant} and vanilla. More cases are in Supplementary.}
  \label{fig:saliency_map}
  \vspace{-0.1in}
\end{figure}

For example, the first case in Figure \ref{fig:saliency_map} is of a mutagenic molecule in that it has two methyl groups attached to oxygen atoms, which are vulnerable to radical attack. Attacked by radical, the attached methyl groups are transferred to DNA base, so-called DNA methylation, entailing the change of DNA base property, resulting in mutation. While the model trained with \emph{Graph Transplant} attends to the appropriate region colored by red, representing high node saliency, the vanilla model gives an incorrect answer with highlighting the unrelated spot.  Next, the right molecule has `OH' and `COOH' functional groups which can be ionized, the resulting exposed electron could attack DNA base. The saliency of our method points out `OH' the most. On the other hand, the vanilla model focuses on the core of symmetric structure locally and considers it stable so that it is non-mutagenic. 

\vspace{-0.08in}
\section{Conclusion}

We proposed \emph{Graph Transplant}, the first input-level Mixup strategy for graph-structured data, that generates mixed graphs via the guidance of the node saliency. To create semantically valuable graphs, \emph{Graph Transplant} exploits the salient subgraphs as mixing units while preserving the local structure. Furthermore, supervisions for augmented graphs are determined in a context-aware manner to rectify the data-label mismatching problem. Through the extensive experiments, \emph{Graph Transplant} demonstrated its effectiveness in that it outperforms the existing graph augmentations across the multiple domains of graphs and GNN architectures. 

\section{Acknowledgements}
This work was supported by Institute of Information \& communications Technology Planning \& Evaluation (IITP) grant funded by the Korea government(MSIT) (No.2019-0-00075, Artificial Intelligence Graduate School Program(KAIST), No.2019-0-00075, Artificial Intelligence Innovation Hub), and Samsung Electronics Co., Ltd (No.IO201214-08133-01).

\bibliography{main}

\begin{thebibliography}{47}
\providecommand{\natexlab}[1]{#1}

\bibitem[{Bengio, Courville, and Vincent(2013)}]{bengio2013representation}
Bengio, Y.; Courville, A.; and Vincent, P. 2013.
\newblock Representation learning: A review and new perspectives.
\newblock \emph{IEEE transactions on pattern analysis and machine
  intelligence}, 35(8): 1798--1828.

\bibitem[{Chen et~al.(2020)Chen, Lin, Li, Li, Zhou, and Sun}]{adaedge}
Chen, D.; Lin, Y.; Li, W.; Li, P.; Zhou, J.; and Sun, X. 2020.
\newblock Measuring and relieving the over-smoothing problem for graph neural
  networks from the topological view.
\newblock In \emph{Proceedings of the AAAI Conference on Artificial
  Intelligence}, volume~34, 3438--3445.

\bibitem[{Dai et~al.(2018)Dai, Li, Tian, Huang, Wang, Zhu, and
  Song}]{gradargmax}
Dai, H.; Li, H.; Tian, T.; Huang, X.; Wang, L.; Zhu, J.; and Song, L. 2018.
\newblock Adversarial attack on graph structured data.
\newblock In \emph{International conference on machine learning}, 1115--1124.
  PMLR.

\bibitem[{Devries and Taylor(2017)}]{cutout}
Devries, T.; and Taylor, G.~W. 2017.
\newblock Improved Regularization of Convolutional Neural Networks with Cutout.
\newblock \emph{CoRR}.

\bibitem[{Errica et~al.(2019)Errica, Podda, Bacciu, and Micheli}]{fair}
Errica, F.; Podda, M.; Bacciu, D.; and Micheli, A. 2019.
\newblock A fair comparison of graph neural networks for graph classification.
\newblock \emph{arXiv preprint arXiv:1912.09893}.

\bibitem[{Gilmer et~al.(2017)Gilmer, Schoenholz, Riley, Vinyals, and
  Dahl}]{neural_message_passing}
Gilmer, J.; Schoenholz, S.~S.; Riley, P.~F.; Vinyals, O.; and Dahl, G.~E. 2017.
\newblock Neural message passing for quantum chemistry.
\newblock In \emph{International Conference on Machine Learning}, 1263--1272.
  PMLR.

\bibitem[{Guo, Mao, and Zhang(2019{\natexlab{a}})}]{nlpmixup}
Guo, H.; Mao, Y.; and Zhang, R. 2019{\natexlab{a}}.
\newblock Augmenting data with mixup for sentence classification: An empirical
  study.
\newblock \emph{arXiv preprint arXiv:1905.08941}.

\bibitem[{Guo, Mao, and Zhang(2019{\natexlab{b}})}]{adamixup}
Guo, H.; Mao, Y.; and Zhang, R. 2019{\natexlab{b}}.
\newblock Mixup as locally linear out-of-manifold regularization.
\newblock In \emph{Proceedings of the AAAI Conference on Artificial
  Intelligence}, volume~33, 3714--3722.

\bibitem[{Hamilton, Ying, and Leskovec(2017)}]{hamilton2017inductive}
Hamilton, W.~L.; Ying, R.; and Leskovec, J. 2017.
\newblock Inductive representation learning on large graphs.
\newblock \emph{arXiv preprint arXiv:1706.02216}.

\bibitem[{He et~al.(2016)He, Zhang, Ren, and Sun}]{resnet}
He, K.; Zhang, X.; Ren, S.; and Sun, J. 2016.
\newblock Deep residual learning for image recognition.
\newblock In \emph{Proceedings of the IEEE conference on computer vision and
  pattern recognition}, 770--778.

\bibitem[{Hendrycks et~al.(2019)Hendrycks, Mu, Cubuk, Zoph, Gilmer, and
  Lakshminarayanan}]{augmix}
Hendrycks, D.; Mu, N.; Cubuk, E.~D.; Zoph, B.; Gilmer, J.; and
  Lakshminarayanan, B. 2019.
\newblock Augmix: A simple data processing method to improve robustness and
  uncertainty.
\newblock \emph{arXiv preprint arXiv:1912.02781}.

\bibitem[{Hu et~al.(2020)Hu, Fey, Zitnik, Dong, Ren, Liu, Catasta, and
  Leskovec}]{hu2020ogb}
Hu, W.; Fey, M.; Zitnik, M.; Dong, Y.; Ren, H.; Liu, B.; Catasta, M.; and
  Leskovec, J. 2020.
\newblock Open Graph Benchmark: Datasets for Machine Learning on Graphs.
\newblock \emph{arXiv preprint arXiv:2005.00687}.

\bibitem[{Jang, Gu, and Poole(2016)}]{gumbel-softmax}
Jang, E.; Gu, S.; and Poole, B. 2016.
\newblock Categorical reparameterization with gumbel-softmax.
\newblock \emph{arXiv preprint arXiv:1611.01144}.

\bibitem[{Kazius, McGuire, and Bursi(2005)}]{mutagen}
Kazius, J.; McGuire, R.; and Bursi, R. 2005.
\newblock Derivation and validation of toxicophores for mutagenicity
  prediction.
\newblock \emph{Journal of medicinal chemistry}, 48(1): 312--320.

\bibitem[{Kim et~al.(2021)Kim, Choo, Jeong, and Song}]{comixup}
Kim, J.-H.; Choo, W.; Jeong, H.; and Song, H.~O. 2021.
\newblock Co-mixup: Saliency guided joint mixup with supermodular diversity.
\newblock \emph{arXiv preprint arXiv:2102.03065}.

\bibitem[{Kim, Choo, and Song(2020)}]{puzzlemix}
Kim, J.-H.; Choo, W.; and Song, H.~O. 2020.
\newblock Puzzle mix: Exploiting saliency and local statistics for optimal
  mixup.
\newblock In \emph{International Conference on Machine Learning}, 5275--5285.
  PMLR.

\bibitem[{Kipf and Welling(2016)}]{gcnn}
Kipf, T.~N.; and Welling, M. 2016.
\newblock Semi-supervised classification with graph convolutional networks.
\newblock \emph{arXiv preprint arXiv:1609.02907}.

\bibitem[{Luch(2005)}]{luch2005nature}
Luch, A. 2005.
\newblock Nature and nurture--lessons from chemical carcinogenesis.
\newblock \emph{Nature Reviews Cancer}, 5(2): 113--125.

\bibitem[{Medennikov et~al.(2018)Medennikov, Khokhlov, Romanenko, Popov,
  Tomashenko, Sorokin, and Zatvornitskiy}]{medennikov2018investigation}
Medennikov, I.; Khokhlov, Y.~Y.; Romanenko, A.; Popov, D.; Tomashenko, N.~A.;
  Sorokin, I.; and Zatvornitskiy, A. 2018.
\newblock An Investigation of Mixup Training Strategies for Acoustic Models in
  ASR.
\newblock In \emph{INTERSPEECH}, 2903--2907.

\bibitem[{Meng and Zhang(2019)}]{isonn}
Meng, L.; and Zhang, J. 2019.
\newblock Isonn: Isomorphic neural network for graph representation learning
  and classification.
\newblock \emph{arXiv preprint arXiv:1907.09495}.

\bibitem[{Neil et~al.(2018)Neil, Briody, Lacoste, Sim, Creed, and
  Saffari}]{neil2018interpretable}
Neil, D.; Briody, J.; Lacoste, A.; Sim, A.; Creed, P.; and Saffari, A. 2018.
\newblock Interpretable graph convolutional neural networks for inference on
  noisy knowledge graphs.
\newblock \emph{arXiv preprint arXiv:1812.00279}.

\bibitem[{Pang, Xu, and Zhu(2019)}]{adv_mixup}
Pang, T.; Xu, K.; and Zhu, J. 2019.
\newblock Mixup inference: Better exploiting mixup to defend adversarial
  attacks.
\newblock \emph{arXiv preprint arXiv:1909.11515}.

\bibitem[{Riesen and Bunke(2008)}]{coil-del}
Riesen, K.; and Bunke, H. 2008.
\newblock IAM graph database repository for graph based pattern recognition and
  machine learning.
\newblock In \emph{Joint IAPR International Workshops on Statistical Techniques
  in Pattern Recognition (SPR) and Structural and Syntactic Pattern Recognition
  (SSPR)}, 287--297. Springer.

\bibitem[{Rong et~al.(2019)Rong, Huang, Xu, and Huang}]{rong2019dropedge}
Rong, Y.; Huang, W.; Xu, T.; and Huang, J. 2019.
\newblock Dropedge: Towards deep graph convolutional networks on node
  classification.
\newblock \emph{arXiv preprint arXiv:1907.10903}.

\bibitem[{Schomburg et~al.(2004)Schomburg, Chang, Ebeling, Gremse, Heldt, Huhn,
  and Schomburg}]{enzymes}
Schomburg, I.; Chang, A.; Ebeling, C.; Gremse, M.; Heldt, C.; Huhn, G.; and
  Schomburg, D. 2004.
\newblock BRENDA, the enzyme database: updates and major new developments.
\newblock \emph{Nucleic acids research}, 32(suppl\_1): D431--D433.

\bibitem[{Selvaraju et~al.(2017)Selvaraju, Cogswell, Das, Vedantam, Parikh, and
  Batra}]{grad-cam}
Selvaraju, R.~R.; Cogswell, M.; Das, A.; Vedantam, R.; Parikh, D.; and Batra,
  D. 2017.
\newblock Grad-cam: Visual explanations from deep networks via gradient-based
  localization.
\newblock In \emph{Proceedings of the IEEE international conference on computer
  vision}, 618--626.

\bibitem[{Shrikumar, Greenside, and Kundaje(2017)}]{saliency_activation}
Shrikumar, A.; Greenside, P.; and Kundaje, A. 2017.
\newblock Learning important features through propagating activation
  differences.
\newblock In \emph{International Conference on Machine Learning}, 3145--3153.
  PMLR.

\bibitem[{Simonovsky and Komodakis(2017)}]{ecc}
Simonovsky, M.; and Komodakis, N. 2017.
\newblock Dynamic edge-conditioned filters in convolutional neural networks on
  graphs.
\newblock In \emph{Proceedings of the IEEE conference on computer vision and
  pattern recognition}, 3693--3702.

\bibitem[{Simonyan, Vedaldi, and Zisserman(2013)}]{simonyan}
Simonyan, K.; Vedaldi, A.; and Zisserman, A. 2013.
\newblock Deep inside convolutional networks: Visualising image classification
  models and saliency maps.
\newblock \emph{arXiv preprint arXiv:1312.6034}.

\bibitem[{Srivastava et~al.(2014)Srivastava, Hinton, Krizhevsky, Sutskever, and
  Salakhutdinov}]{srivastava2014dropout}
Srivastava, N.; Hinton, G.; Krizhevsky, A.; Sutskever, I.; and Salakhutdinov,
  R. 2014.
\newblock Dropout: a simple way to prevent neural networks from overfitting.
\newblock \emph{The journal of machine learning research}, 15(1): 1929--1958.

\bibitem[{Vaswani et~al.(2017)Vaswani, Shazeer, Parmar, Uszkoreit, Jones,
  Gomez, Kaiser, and Polosukhin}]{transformer}
Vaswani, A.; Shazeer, N.; Parmar, N.; Uszkoreit, J.; Jones, L.; Gomez, A.~N.;
  Kaiser, L.; and Polosukhin, I. 2017.
\newblock Attention is all you need.
\newblock \emph{arXiv preprint arXiv:1706.03762}.

\bibitem[{Veli{\v{c}}kovi{\'c} et~al.(2017)Veli{\v{c}}kovi{\'c}, Cucurull,
  Casanova, Romero, Lio, and Bengio}]{gatt}
Veli{\v{c}}kovi{\'c}, P.; Cucurull, G.; Casanova, A.; Romero, A.; Lio, P.; and
  Bengio, Y. 2017.
\newblock Graph attention networks.
\newblock \emph{arXiv preprint arXiv:1710.10903}.

\bibitem[{Verma et~al.(2019{\natexlab{a}})Verma, Lamb, Beckham, Najafi,
  Mitliagkas, Lopez-Paz, and Bengio}]{manifold_mixup}
Verma, V.; Lamb, A.; Beckham, C.; Najafi, A.; Mitliagkas, I.; Lopez-Paz, D.;
  and Bengio, Y. 2019{\natexlab{a}}.
\newblock Manifold mixup: Better representations by interpolating hidden
  states.
\newblock In \emph{International Conference on Machine Learning}, 6438--6447.
  PMLR.

\bibitem[{Verma et~al.(2019{\natexlab{b}})Verma, Qu, Lamb, Bengio, Kannala, and
  Tang}]{graphmix}
Verma, V.; Qu, M.; Lamb, A.; Bengio, Y.; Kannala, J.; and Tang, J.
  2019{\natexlab{b}}.
\newblock Graphmix: Regularized training of graph neural networks for
  semi-supervised learning.
\newblock \emph{arXiv preprint arXiv:1909.11715}.

\bibitem[{Wale, Watson, and Karypis(2008)}]{nci1}
Wale, N.; Watson, I.~A.; and Karypis, G. 2008.
\newblock Comparison of descriptor spaces for chemical compound retrieval and
  classification.
\newblock \emph{Knowledge and Information Systems}, 14(3): 347--375.

\bibitem[{Wang et~al.(2021)Wang, Wang, Liang, Cai, and Hooi}]{wang2021mixup}
Wang, Y.; Wang, W.; Liang, Y.; Cai, Y.; and Hooi, B. 2021.
\newblock Mixup for Node and Graph Classification.
\newblock In \emph{Proceedings of the Web Conference 2021}, 3663--3674.

\bibitem[{Xu et~al.(2018)Xu, Hu, Leskovec, and Jegelka}]{ginn}
Xu, K.; Hu, W.; Leskovec, J.; and Jegelka, S. 2018.
\newblock How powerful are graph neural networks?
\newblock \emph{arXiv preprint arXiv:1810.00826}.

\bibitem[{Yan et~al.(2008)Yan, Cheng, Han, and Yu}]{large_mol}
Yan, X.; Cheng, H.; Han, J.; and Yu, P.~S. 2008.
\newblock Mining Significant Graph Patterns by Leap Search.
\newblock In \emph{Proceedings of the 2008 ACM SIGMOD International Conference
  on Management of Data}, SIGMOD '08, 433–444. Association for Computing
  Machinery.

\bibitem[{Yanardag and Vishwanathan(2015)}]{socialdataset}
Yanardag, P.; and Vishwanathan, S. 2015.
\newblock Deep graph kernels.
\newblock In \emph{Proceedings of the 21th ACM SIGKDD international conference
  on knowledge discovery and data mining}, 1365--1374.

\bibitem[{Ying et~al.(2019)Ying, Bourgeois, You, Zitnik, and
  Leskovec}]{gnnexplainer}
Ying, R.; Bourgeois, D.; You, J.; Zitnik, M.; and Leskovec, J. 2019.
\newblock Gnnexplainer: Generating explanations for graph neural networks.
\newblock \emph{Advances in neural information processing systems}, 32: 9240.

\bibitem[{Ying et~al.(2018)Ying, You, Morris, Ren, Hamilton, and
  Leskovec}]{diffpool}
Ying, R.; You, J.; Morris, C.; Ren, X.; Hamilton, W.~L.; and Leskovec, J. 2018.
\newblock Hierarchical graph representation learning with differentiable
  pooling.
\newblock \emph{arXiv preprint arXiv:1806.08804}.

\bibitem[{You et~al.(2020)You, Chen, Sui, Chen, Wang, and Shen}]{graphcl}
You, Y.; Chen, T.; Sui, Y.; Chen, T.; Wang, Z.; and Shen, Y. 2020.
\newblock Graph contrastive learning with augmentations.
\newblock \emph{Advances in Neural Information Processing Systems}, 33.

\bibitem[{Yun et~al.(2019)Yun, Han, Oh, Chun, Choe, and Yoo}]{cutmix}
Yun, S.; Han, D.; Oh, S.~J.; Chun, S.; Choe, J.; and Yoo, Y. 2019.
\newblock Cutmix: Regularization strategy to train strong classifiers with
  localizable features.
\newblock In \emph{Proceedings of the IEEE/CVF International Conference on
  Computer Vision}, 6023--6032.

\bibitem[{Zhang et~al.(2017)Zhang, Cisse, Dauphin, and Lopez-Paz}]{mixup}
Zhang, H.; Cisse, M.; Dauphin, Y.~N.; and Lopez-Paz, D. 2017.
\newblock mixup: Beyond empirical risk minimization.
\newblock \emph{arXiv preprint arXiv:1710.09412}.

\bibitem[{Zhao et~al.(2020)Zhao, Liu, Neves, Woodford, Jiang, and Shah}]{gaug}
Zhao, T.; Liu, Y.; Neves, L.; Woodford, O.; Jiang, M.; and Shah, N. 2020.
\newblock Data Augmentation for Graph Neural Networks.
\newblock \emph{arXiv preprint arXiv:2006.06830}.

\bibitem[{Zhou et~al.(2016)Zhou, Khosla, Lapedriza, Oliva, and
  Torralba}]{zhou2016learning}
Zhou, B.; Khosla, A.; Lapedriza, A.; Oliva, A.; and Torralba, A. 2016.
\newblock Learning deep features for discriminative localization.
\newblock In \emph{Proceedings of the IEEE conference on computer vision and
  pattern recognition}, 2921--2929.

\bibitem[{Zhou, Shen, and Xuan(2020)}]{augforgraphclf}
Zhou, J.; Shen, J.; and Xuan, Q. 2020.
\newblock Data Augmentation for Graph Classification.
\newblock In \emph{Proceedings of the 29th ACM International Conference on
  Information \& Knowledge Management}, 2341--2344.

\end{thebibliography}

\clearpage
\appendix
\section{Results of GCN}
We report the results of GCN in Table \ref{tb:gcn_main} which have been omitted in Table \ref{tb:maintable} in the main paper due to the space constraint. As in the Table \ref{tb:maintable}, \emph{Graph Transplant} consistently shows its superiority compared to the other baselines.

\begin{table*}[t]
\caption{\small To compare our method `$\emph{Graph Transplant}_{DP/EP}$' with the other baselines, we report the averaged accuracy and the standard deviation of 3 repeats of 5-fold cross validation on five graph classification benchmark datasets for GCN architecture. 
Note that MaskN can not be applied to COLLAB which does not have node features.} 
\vspace{-0.15in}
\begin{center}
\setlength{\tabcolsep}{6.0pt} 
\begin{scriptsize}
\begin{tabular}{@{\extracolsep{8pt}}rlccccc@{}}
\toprule
& \textbf{Method} & COLLAB & NCI1 & ENZYMES & Mutagenicity & COIL-DEL \\
\cline{2-7}
\rule{0pt}{2.5ex}  
\multirow{9}{*}{\rotatebox{90}{GCN}} & Vanilla & 0.809 \tiny{$\pm 0.008$} & 0.742 \tiny{$\pm 0.015$} 
& 0.606 \tiny{$\pm 0.047$} & 0.809 \tiny{$\pm 0.006$}& 0.539 \tiny{$\pm 0.015$} \\
                    \cdashline{2-7}
                    \rule{0pt}{2ex}
                    & Manifold Mixup & 0.815 \tiny{$\pm 0.007$} & 0.762 \tiny{$\pm 0.017$} 
                     & 0.633 \tiny{$\pm 0.049$}  & 0.810 \tiny{$\pm 0.013$} &  0.559 \tiny{$\pm 0.019$}\\
                     & M-Evolve &  0.816 \tiny{$\pm 0.005 $} & 0.730 \tiny{$\pm 0.019$} 
                      & 0.615 \tiny{$\pm 0.042$}  & 0.806 \tiny{$\pm 0.011$} &  0.527 \tiny{$\pm 0.014$}\\
                     & DropN & 0.815 \tiny{$\pm 0.009$} & 0.736 \tiny{$\pm 0.024$}   
                     & 0.629 \tiny{$\pm 0.039$} & 0.811 \tiny{$\pm 0.012$} & 0.559 \tiny{$\pm 0.036$}\\
                     & PermE & 0.810 \tiny{$\pm 0.009$} & 0.739 \tiny{$\pm 0.024$} 
                     & 0.623 \tiny{$\pm 0.039$} & 0.814 \tiny{$\pm 0.010$} & 0.562 \tiny{$\pm 0.030$} \\
                     & MaskN & -& 0.731 \tiny{$\pm 0.020$} 
                     & 0.636 \tiny{$\pm 0.034$}& 0.806 \tiny{$\pm 0.011$} & 0.528 \tiny{$\pm 0.025$} \\
                     & SubG & 0.813 \tiny{$\pm 0.010$}  & 0.741 \tiny{$\pm 0.028$}  
                     & 0.623 \tiny{$\pm 0.035$} & 0.808 \tiny{$\pm 0.013$} & 0.551 \tiny{$\pm 0.032$} \\
                     
                     \cline{2-7}
                     \rule{0pt}{2ex}  
                        & \textbf{$\text{Graph Transplant}_{DP}$} & 0.818 \tiny{$\pm 0.009$} & 0.766 \tiny{$\pm 0.019$} 
                     & 0.637 \tiny{$\pm 0.040$} & \textbf{0.818} \tiny{$\pm 0.009$} & 0.565 \tiny{$\pm 0.028$} \\
                     
                     & \textbf{$\text{Graph Transplant}_{EP}$} & \BF{0.822} \tiny{$\pm 0.010$} & \BF{0.770} \tiny{$\pm 0.019$} 
                     & \BF{0.647} \tiny{$\pm 0.044$} & \BF{0.817} \tiny{$\pm 0.011$} & \BF{0.582} \tiny{$\pm 0.017$} \\
\bottomrule
\end{tabular}
\vspace{-0.18 in}
\end{scriptsize}
\end{center}
\label{tb:gcn_main}
\end{table*}

\section{Ablation study}
\begin{figure*}[h]
  \centering
 \begin{minipage}[b]{0.32\textwidth}
    \includegraphics[width=\textwidth]{fig/ablation_NCI1.png}
  \end{minipage}
  \begin{minipage}[b]{0.32\textwidth}
    \includegraphics[width=\textwidth]{fig/abalation_mutagenicity.png}
  \end{minipage}
  \begin{minipage}[b]{0.315\textwidth}
    \includegraphics[width=\textwidth]{fig/ablation_ENZYMES.png}
  \end{minipage}
 \begin{minipage}[b]{0.32\textwidth}
    \includegraphics[width=\textwidth]{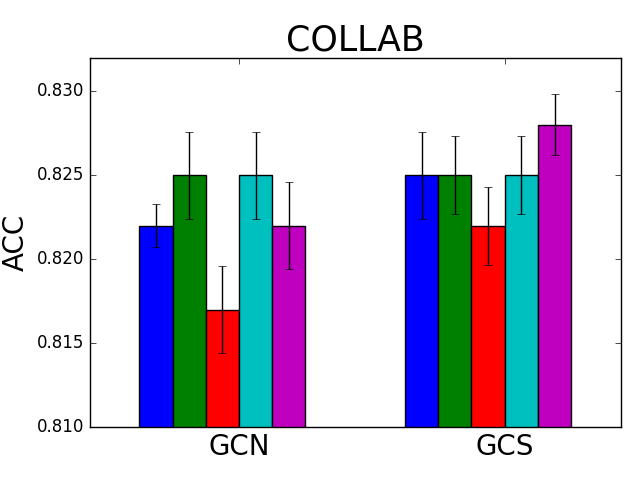}
  \end{minipage}
 \begin{minipage}[b]{0.32\textwidth}
    \includegraphics[width=\textwidth]{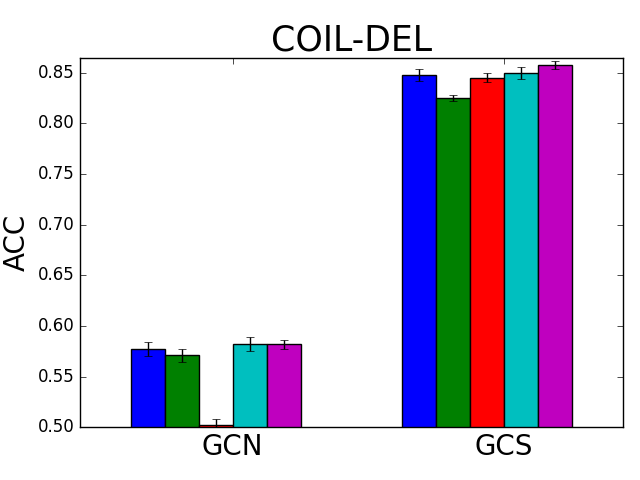}
  \end{minipage}
  \vspace{-0.1in}
  \caption{\small Ablation study. We draw a bar chart for each dataset, where bars are grouped by model architectures within a chart. The different ablation studies are distinguishable by colors. The rightmost bars are of our complete method.}
  \label{fig:ablation_supp}
\end{figure*}

We provide the results of ablation studies on all small datasets less than 10k graph instances including COLLAB and COIL-DEL that are omitted in Section \ref{subsec:ablation}. The consistent tendency can be confirmed in Figure~\ref{fig:ablation_supp} 
that the absence of only one component of \emph{Graph Transplant} often leads to degrading the classification performance. We describe the details of each ablation study in the following paragraphs. 

\paragraph{Edge prediction} 
\emph{Graph Transplant} connects the two subgraphs from the source and the destination graph, respectively, with the edge prediction module $\psi$. To justify the need for this module, we leave the subgraphs disconnected for the newly mixed data. We can of course obtain performance gain without further connection because on the one hand graphs are still being mixed by the readout function $\gamma$ of GNN. However, the results are better with edge prediction as shown in Figure~\ref{fig:ablation}, where the first bar labeled by `w/o Edge Prediction' and the last bar for our `Graph Transplant' is compared for each group.

\paragraph{Label mixing strategies : Size-based vs Saliency-based}
We leverage the node saliency to construct the new labels for the mixed graphs. 
Another intuitive option is to set labels based on the size of subgraphs or the number of nodes. Formally, the label mixing ratio $\lambda$ is defined as $\frac{|\bar{V}_\pi| / |V_\pi|}{|\bar{V}_\pi|/|V_\pi| + |V\setminus \bar{V}|/|V|}$. However, size-based labels can give erroneous signals when the hints about the class are concentrated only in local areas of the graph. In this case, we have to put more weight on the core part. The results from the ablation study on the label construction rule support our argument (see `Size-based Label' in Figure~\ref{fig:ablation}). \emph{Graph Transplant} with salient-based labels has better performances than that of the method with size-based labels.
\paragraph{Comparison of mixing units}
In Section~\ref{subsec:graphmixing}, we adopt the $K$-hop subgraph as a mixing unit. To demonstrate the importance of preserving locality 
when mixing graphs, we compare the performance by setting the scattered nodes as the mixing unit. For fair comparison, we select $N$ salient nodes among the most salient $\min(2N, |V|)$ nodes, where $N$ is the average number of nodes in the partial $K$-hop subgraphs ($N = \mathbb{E}[\sum_{i}^{\mathcal{B}}{|\bar{V}_{\pi_i}}|]$). Results of `Scattered NodeMix' in Figure~\ref{fig:ablation} 
confirm that transplanting the scattered nodes significantly degrades the performance. This implies that mixing dispersive nodes is insufficient to capture the semantics of the original graph. In line with the recent Mixup approaches of the CV ~\cite{cutmix}, the locality of the graph is also a critical factor to avoid generating the noisy samples.

\paragraph{Importance of saliency-based subgraph}
We discussed the significance of exploiting node saliency for extracting the subgraph (Section~\ref{subsec:nodesaliency}). Some practitioners may argue that generating numerous data based on randomness is more important than considering the context of data. To verify the effectiveness of the salient subgraph selection, we conduct comparative experiments with random subgraph selection. The experimental results (`Random Subgraph') in Figure~\ref{fig:ablation}
exhibit that transplanting the salient subgraphs to the destination graph outperforms the random subgraphs selection. The results demonstrate that the guidance of node saliency is beneficial for generating more refined virtual samples.

\section{Additional Experiments}

\subsection{Adversarial Attack}
To verify the robustness of the model trained with \emph{GraphTransplant} toward the adversarial attack, we apply the white-box attack \textit{GradArgmax}~\cite{gradargmax} to the test dataset of each fold. This adversarial attack uses gradients of the loss function with respect to the adjacency matrix. \textit{GradArgmax} determines the $m$ node pairs with the largest magnitude of gradients by a greedy algorithm and adds edges between selected node pairs if the gradient is positive, or removes edges if the gradient is negative. We set $m$ to be 20\% of the average number of edges in each dataset, which is the number of edges to be modified in original graphs. Also, the maximum distance of an edge that can be modified is limited to 4-hop ($b=4$). We reported the averaged accuracy for 5-fold cross-validation. In Table~\ref{supp:adversarial_gin}, our method shows consistent superior performance in another architecture that is omitted in main paper. 

\begin{table*}[h]
 \caption{\footnotesize Evaluation of the model robustness under the adversarial attack. We apply a white-box attack \textit{GradArgmax} to test set of each fold. Averaged accuracies of 5-fold cross-validation are reported.}

\setlength{\columnsep}{1pt}
\begin{center}
\begin{scriptsize}
\setlength{\tabcolsep}{0.8pt} 
\renewcommand{\arraystretch}{1} 
\begin{tabular}{@{\extracolsep{1pt}}lccc@{}}
\toprule
\textbf{Method} & \multicolumn{3}{c}{\textbf{\textit{GradArgmax}} \scriptsize(White-Box Attack) }  \\ 
\cline{2-4}
\textbf{(GIN)} & ENZYMES & Mutagenicity & NCI1  \\
\cline{1-4}
                     Vanilla & 0.480 \tiny{$\pm 0.040$} & 0.595 \tiny{$\pm 0.039$}& 0.643 \tiny{$\pm 0.018$}\\
                     \cdashline{1-4}
                     Manifold Mixup & 0.547 \tiny{$\pm 0.031$}  & 0.568 \tiny{$\pm 0.016$} & 0.660 \tiny{$\pm 0.043$}\\
                     DropN & 0.527 \tiny{$\pm 0.042$} &  0.677 \tiny{$\pm 0.018$}& 0.670 \tiny{$\pm 0.059$} \\
                     PermE & 0.542 \tiny{$\pm 0.023$} & 0.645 \tiny{$\pm 0.016$}  & 0.675 \tiny{$\pm 0.041$}     \\
                     MaskN & 0.548 \tiny{$\pm 0.018$}  & 0.624 \tiny{$\pm 0.063$} & 0.667 \tiny{$\pm 0.034$}       \\
                     SubG &  0.533 \tiny{$\pm 0.031$} & 0.646 \tiny{$\pm 0.009$} & 0.648 \tiny{$\pm 0.020$}       \\
                     \cline{1-4}
                     $\text{\textbf{Graph Transplant}}_{EP}$ & \BF{0.603} \tiny{$\pm 0.040$} & \BF{0.698} \tiny{$\pm 0.031$} & \BF{0.721} \tiny{$\pm 0.024$}   \\
\bottomrule
\end{tabular}
\end{scriptsize}
\end{center}
\label{supp:adversarial_gin}
\end{table*}

\subsection{Accuracy of edge predictor}~\label{subsec:acc_ep}
\begin{figure*}[h]

 \centering
 \includegraphics[width=1\linewidth]{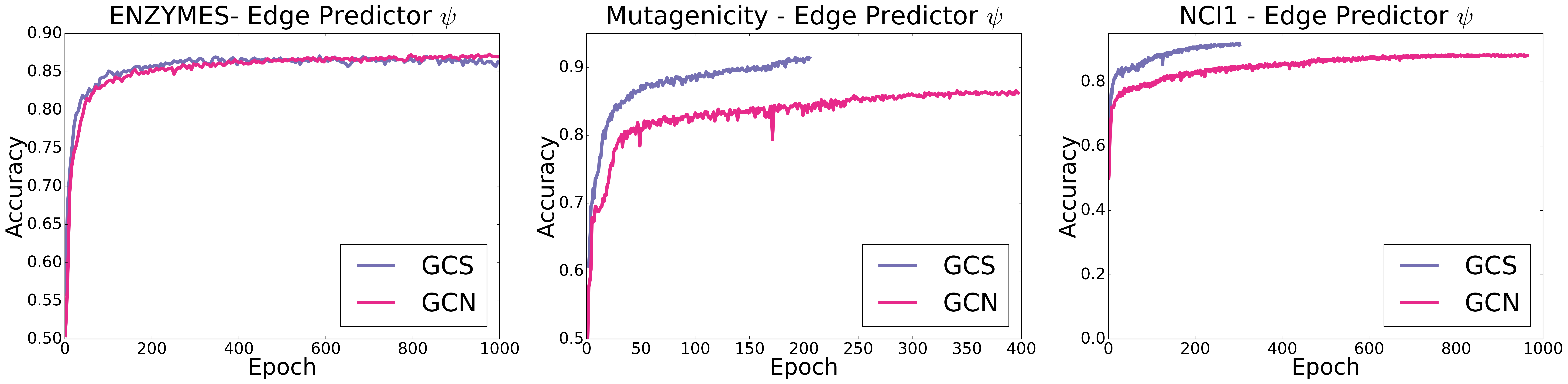}
 \caption{\small Edge prediction accuracy of edge predictor $\psi$ during the training phase. The accuracy was evaluated on the set of node pairs in test graphs. The experiments were conducted on three graph classification benchmark datasets with two GNN architectures - GCN, GCS.}
 \label{fig:ep_curve}
 \end{figure*}

As a preliminary experiment, we demonstrate the accuracy of the edge predictor $\psi$ on the set of node pairs included in test graphs while the model is being trained (Figure \ref{fig:ep_curve}). We observe that the accuracies converge in quite an early phase of training and the performances are satisfactory to be used for the data augmentation purpose that allows some noise.  

\subsection{Sensitivity to different readout functions}\label{subsec:readout}
\begin{figure*}[h]
  \centering
  \begin{minipage}[b]{0.47\textwidth}
    \includegraphics[width=\textwidth]{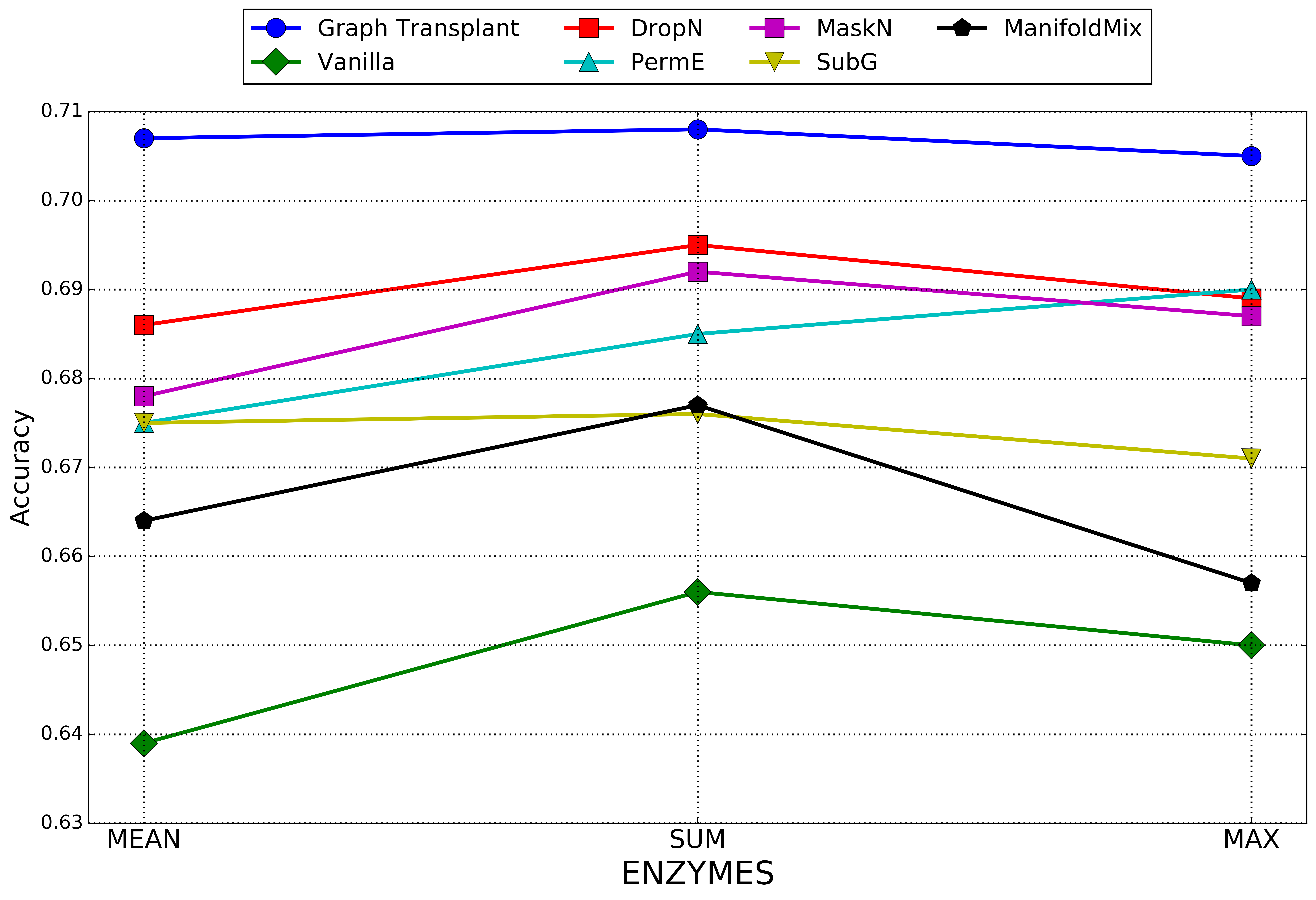}
  \end{minipage}
  \begin{minipage}[b]{0.47\textwidth}
    \includegraphics[width=\textwidth]{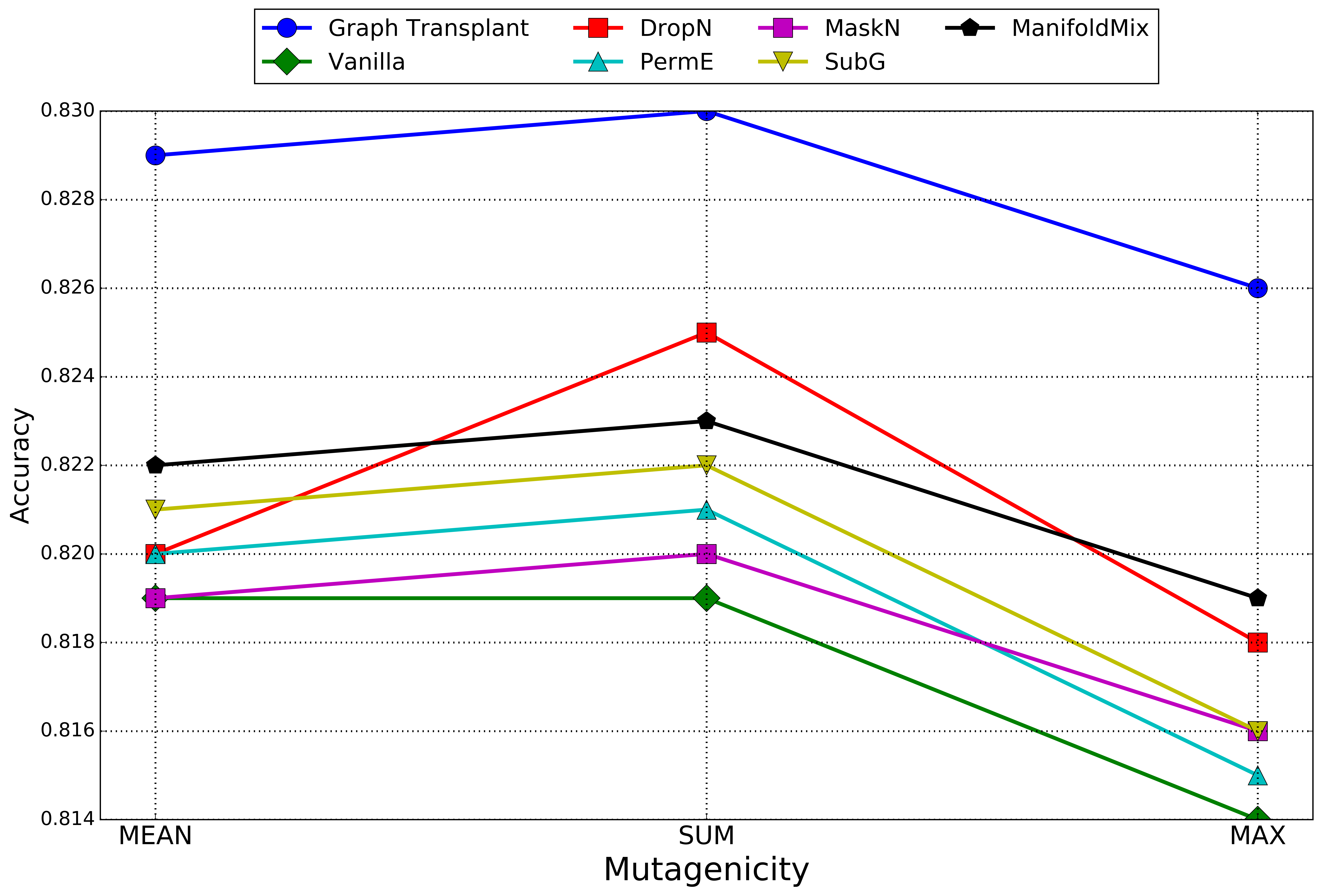}
  \end{minipage}
  \caption{\small Sensitivity to the readout functions (MEAN/SUM/MAX) of \emph{GraphTransplant} and the other baselines. \emph{GraphGrafing} shows consistent performance improvement without dependence on the readout functions. Experiments were conducted on two graph classification benchmark datasets with GCS.}
  \label{fig:readout}
\end{figure*}

We test the sensitivity of our method to readout functions other than mean pooling which is used in main experiments. Our method and the baselines are evaluated for the frequently used readout functions, max pooling and sum pooling. As shown in Figure~\ref{fig:readout}, \emph{Graph Transplant} (\textcolor{blue}{Blue}) outperforms all contenders under the other readout functions.  

\section{Detailed Experimental Settings}

In this section, we cover all the details about the experiments we conducted. 

\subsection{Datasets}

We demonstrate the superiority of \emph{Graph Transplant} compared to the baselines on multiple datasets of various sizes and domains. Data statistics and domains are summarized in Table~\ref{tb:datastat}. In this table, $|\mathcal{D}|$, $|V|$, and $|E|$ denote the number of graph instances, the average number of nodes per graph, and the average number of edges per graph, respectively.

As investigated by \cite{fair}, leveraging the node degrees as input features is generally beneficial to improve performances for non-attributed graphs (social datasets). We also adopt the node degrees as input features for COLLAB. Since our method is designed to address graph-structured data with node features and edge weights, we modify ogbg-ppa, which has only edge features but without node features, so that edge features are leveraged to construct node features (mean of connected edges) and then removed.

\begin{table}[h] 
\small
\center
\caption{\small Data statistics of graph classification benchmark datasets.}
\setlength{\tabcolsep}{1.0pt} 
\begin{tabular}{l|l|c|c|c}
\toprule
    Name   & Domain  & $|\mathcal{D}|$ & $|V|$  & $|E|$  \\
    \hline
    COLLAB & Social Networks  & 5000 & 74.5 & 2457.8 \\
    NCI1 & Molecules & 4110 & 29.9 & 28.5 \\
    Mutagenicity & Molecules & 4337 & 30.3& 30.8\\
    ENZYMES & Bioinformatics & 600 & 32.6 & 62.1 \\
    COIL-DEL & Computer Vision & 3900 & 21.5 & 54.2 \\
    NCI-H23 & Molecules& 40353& 26.1& 28.1\\
    MOLT-4 & Molecules &39765&26.1&28.1\\
    P388 & Molecules&41472&22.1&23.6\\
    ogbg-ppa & Bioinformatics &158100&243.4&2266.1\\
    
\bottomrule
\end{tabular}
\label{tb:datastat}
\end{table}

\subsection{Architecture}

We evaluate our augmentation method with four representative architectures of GNNs: GCN, GCN with learnable skip connections (GCS), GAT, and GIN. The following are common settings for the GNNs. We use ReLU as an activation function after every GNN layer. We set the hidden dimension of the GNNs to 128 and tune the number of layers from 3 to 5. To obtain graph representations, we apply the element-wise mean function to the last node representations of each graph. The mean pooling function can be replaced by the max/add pooling layers. We verify that \emph{Graph Transplant} surpasses all the baselines with the other pooling layers as well in Section \ref{subsec:readout}. The graph representations are passed to an MLP classifier with one 128-dimensional hidden layer. Dropout is applied in this MLP classifier with a drop rate of 0.5. The definition of each GNN layer is described in the following paragraphs. 

Let $e_{u, v}$ be the weight of edge $\{u, v\}\in E$, $d_v$ be degree of node $v$ which is the sum of weights $\sum_{u\in\mathcal{N}(v)} e_{u, v}$ of edges connected to $v$. 
\paragraph{GCN \cite{gcnn}} Each $l$-th GCN layer aggregates linearly transformed messages $\Theta^{(l)} x_u^{(l)}$  weighted by $e_{u, v} / \sqrt{\hat{d}_u \hat{d}_v}$ with inserted self-loops. Update of $l$-th GCN layer is conducted as  

\begin{align}
x_v^{(l+1)} = \Theta^{(l)} \sum_{u\in \mathcal{N}(v) \cup \{v\}} \frac{e_{v, u}}{\sqrt{\hat{d}_u \hat{d}_v}} x_u^{(l)} 
\end{align}
where  $\hat{d}_v =  1 + d_v$ and $e_{v, v} = 1$ for all $v$.

\paragraph{GCN with learnable skip connection (GCS)} This layer is similar to GCN but has a learnable skip connection rather than inserting self-loop:

\begin{align}
x_v^{(l+1)} = \Theta_{\texttt{skip}}^{(l)} x_v^{(l)} + \Theta^{(l)} \sum_{u\in \mathcal{N}(v)} \frac{e_{v, u}}{\sqrt{d_u d_v}} x_u^{(l)}
\end{align} 

by introducing additional parameters for the skip connection $\Theta_\texttt{skip}^{(l)}$.

\paragraph{GAT \cite{gatt}} GAT utilizes the attention mechanism when aggregating messages from neighbors as follows.

\begin{align}
x_v^{(l+1)} = \Theta^{(l)} \sum_{u\in \mathcal{N}(u) \cup \{v\}} \alpha_{u, v}^{(l)} x_u^{(l)}, 
\end{align}
where the attention coefficients $\alpha_{u, v}$ are computed as
\begin{align}
    \alpha_{u, v}^{(l)} &= \frac{\mathbbm{1}_{\{e_{u, v} >0\}}\exp(s_{u, v})}{\sum_{w\in \mathcal{N}(v) \cup \{v\}}\mathbbm{1}_{\{e_{w, v} >0\}}\exp(s_{w, v}) }
    \\
    \text{where}\nonumber \\
    s_{u, v} &= \text{LeakyReLU}(\boldsymbol{a}^\top[\Theta^{(l)} x_u^{(l)}||\Theta^{(l)} x_v^{(l)}])\nonumber
\end{align}
with the learnable parameter $\boldsymbol{a}$ and LeakyReLU where the angle of the negative slope is 0.2.

\paragraph{GIN \cite{ginn}}
GIN has more complex architecture to be as powerful as the Weisfeiler-Lehman graph isomorphism test. With an expressive deep model $h_{\Theta^{(l)}}$ such as an MLP, GIN updates node features as
\begin{align}
    x_v^{(l+1)} = h_{\Theta^{(l)}}((1 + \epsilon) x_v^{(l)} + \sum_{u\in\mathcal{N}(v)} e_{u, v} x_u^{(l)} ).
\end{align} 
In our case, we use an MLP with one 128-dimensional hidden layer and the ReLU activation function for $h_\Theta$. 

\paragraph{Edge predictor $\psi$} For the edge predictor, inputs or pairs of node representations are taken from the output of the last GNN layer. We use an MLP with a ReLU activation function and 2 hidden layers with the dimension of [128, 64]. This module outputs the probability of connectivity for the input node pairs. The learning curves of edge predictor are provided in Section~\ref{subsec:acc_ep}. 

\subsection{Implementation details of \emph{Graph Transplant}}
In our experiment, $R$ is set to 10 for small-sized datasets (< 10k) and tuned among $\{5, 10, 15\}$ for three medium-sized molecule datasets ($\sim$ 40k). We use the discrete $K$-hop space $\mathcal{K} = \{1,2,3\}$ for all experiments. To boost the diversity of $K$-hop subgraphs, we stochastically sample the partial subgraph from the full $K$-hop subgraph using the ratio  $p \sim Beta(2, 2)$ (Algorithm~\ref{alg:paritial subgraph}). For the inputs of the edge prediction module and node saliency, we decide to utilize the latent features after the last GNN layer that can capture the highest-level features.

\subsection{Baselines}

In this section, we provide more detailed descriptions of the baselines, their intuitive priors, and the hyperparameter search space. 

\paragraph{Manifold Mixup~\cite{wang2021mixup}}

Since the node feature matrices are compressed into one hidden vector after the readout layer $\gamma$, we can mix these two hidden vectors using the mixing ratio $\lambda$. We sample the mixing ratio $\lambda$ from $Beta(2,2)$ as the Manifold Mixup~\cite{manifold_mixup}.

\paragraph{M-Evolve~\cite{augforgraphclf}} 
This method first perturbs the edges with a heuristically designed open-triad edge swapping method. Then, noisy-augmented graphs are filtered out based on the prediction of validation graphs. We tweak the framework a bit so that augmentation and filtration are conducted for every training step and filtration criterion is updated after each training epoch. We tune the perturbation (add/remove) ratio for \{0.2, 0.4\}.

\paragraph{Node dropping (DropN)}
Similar to Cutout \cite{cutout}, DropN randomly drops a certain portion of nodes. The prior of this approach is that even if the specific portion of the graph is missing, there is no significant damage to the graph property. 
We tune the dropping ratio for \{0.2, 0.4\}.

\paragraph{Edge perturbation (PermE)}
Manipulating the connectivity of the graph is one of the most common strategies for graph augmentation. 
PermE randomly adds and removes a certain ratio of edges. This approach has the intuitive premise that PermE can improve the robustness of the model against the connectivity perturbations.  We tune the perturbation (add/remove) ratio for \{0.2, 0.4\}.

\paragraph{Attribute masking (MaskN)}
In line with Dropout \cite{srivastava2014dropout}, MaskN stochastically masks a certain ratio of node features. The underlying prior of MaskN is that the generalization performance can be enhanced while the network is trained to predict graph semantics with only the remaining features. We tune the masking ratio for \{0.2, 0.4\}.

\paragraph{Subgraph (SubG)}
Using the random walk, SubG extracts the subgraph from the original graph. The prior for SubG is that the local subgraph can represent the semantics of the entire graph. We tune the size of the subgraph (compared to the full-graph) for \{0.6, 0.8\}.

\subsection{Evaluation protocol}

We use Adam optimizer without weight decay, with an initial learning rate of 0.0005 for every network. The learning rate is decayed by the factor of 0.5 if there is no decrease in validation loss for 1000 iterations. For all experiments on small datasets ($<$ 10k), training is conducted for 1000 epochs with a batch size of 128, but a batch size of COLLAB is set to 64 due to memory limitation. Early stopping occurs when there is no improvement of validation accuracy for 1500 iterations. The continuous features of nodes are normalized by their mean and variance (standard normalization) of the training and the validation set.  We slightly change some settings for medium/large-scale datasets. For the medium-sized datasets with about 40k graph instances, patience for early stopping and learning rate decay are set to 50 and 20 epochs, respectively. For the largest ogbg-ppa, these are adjusted to 10 and 5 epochs. Because ogbg-ppa has relatively large graphs with a number of nodes and edges compared to the others, we set the batch size of this dataset to 64. For all experiments, we tune the hyperparameters of baselines and ours to maximize the mean of the validation accuracy. The reported results are the best result by altering the number of layers of GNN from 3 to 5.


\section{Additional qualitative analysis}

\begin{figure*}[h]
  \centering
  \includegraphics[width=0.9\linewidth]{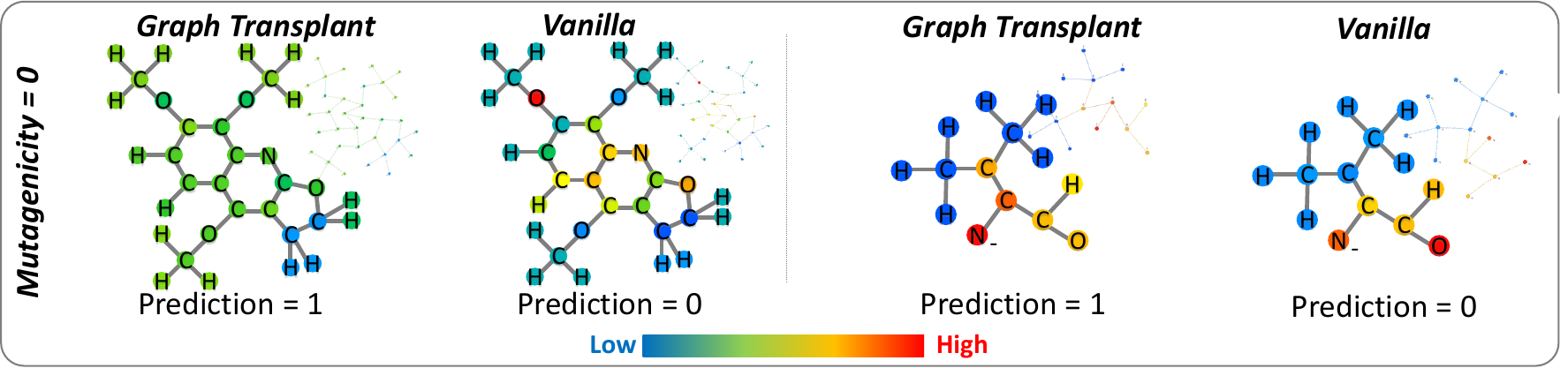}
  \vspace{-0.05in}
  \caption{\small Comparison of saliency maps from the trained models with \emph{Graph Transplant} and the vanilla for two non-mutagenic cases.}
  \label{fig:saliency_map_2}
\end{figure*}

In addition to the examples we present for qualitative analysis in Section~\ref{subsec:analysis}, we include two more cases in this section. 
For the left non-mutagenic example, vanilla training misleads the model to classify it as mutagenic by shortsightedly observing the presence of the oxygen as a shortcut. However, the saliencies of our method are relatively even to capture the overall structure and give a correct answer. In right example, both methods answer correctly 
exhibiting similar saliency maps.

\section{Examples of \emph{Graph Transplant}}

We visualize the results of \emph{Graph Transplant} on multiple datasets. There are three columns. The first and second ones are of the source and the destination graphs. The nodes colored by green/red are selected/remaining nodes of the source/destination graphs with their saliency ratio $\mathcal{I}_{source}/ \mathcal{I}_{destination}$. The last column shows the mixed graphs $G'$ where edges between red and green nodes are the factitious ones generated by the edge prediction module. The label weights $\lambda_{G'}$ for the newly mixed graphs $G'$ are adaptively assigned. The node features are omitted and only the graph structure is represented in the figures.

\begin{figure*}[h]
  
  
  \centering
  \textbf{ENZYMES}\par\medskip
   \begin{minipage}[b]{0.95\textwidth}
    \includegraphics[width=\columnwidth]{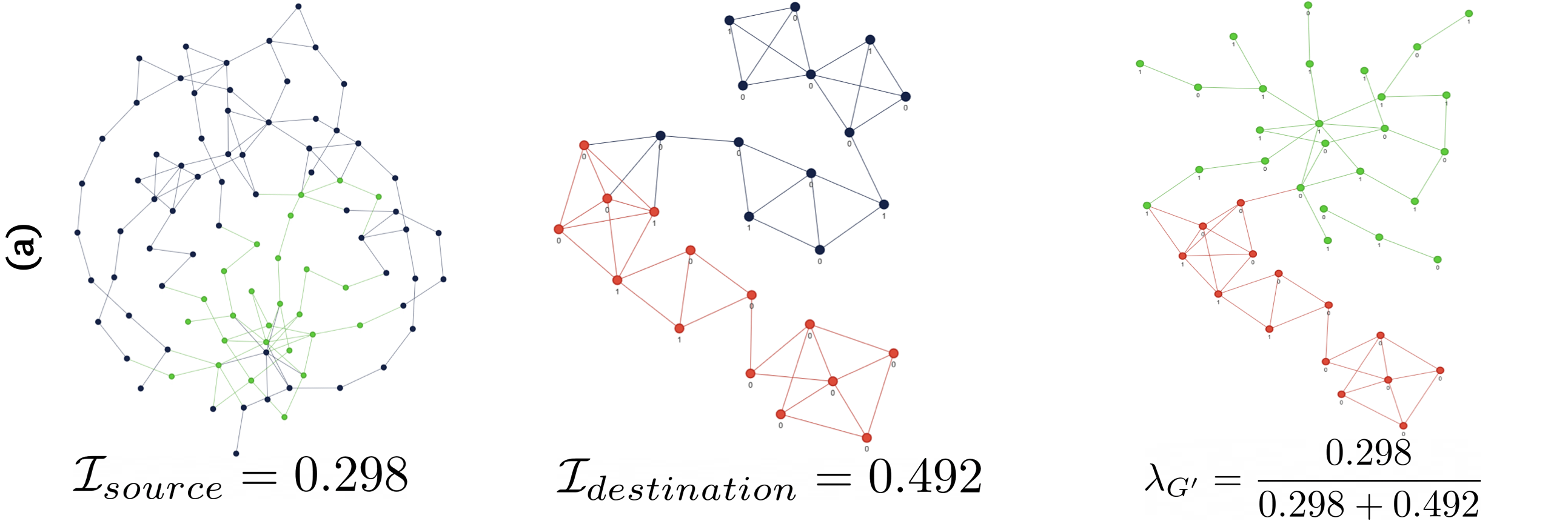}
  \end{minipage}
  
  \vspace{0.7cm}
  
  \begin{minipage}[b]{0.95\textwidth}
    \includegraphics[width=\columnwidth]{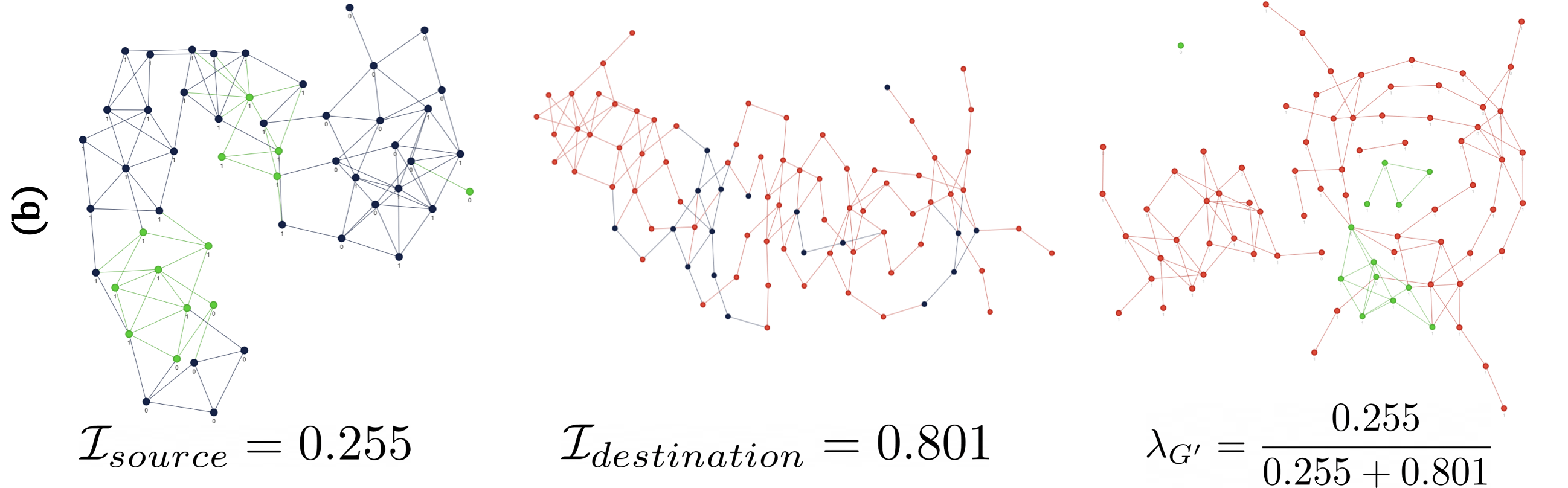}
  \end{minipage}
  
  
  \label{fig:enzymes}
\end{figure*}

\begin{figure*}[h]
  \centering
  \textbf{COIL-DEL}\par\medskip
  \begin{minipage}[b]{0.95\textwidth}
    \includegraphics[width=\columnwidth]{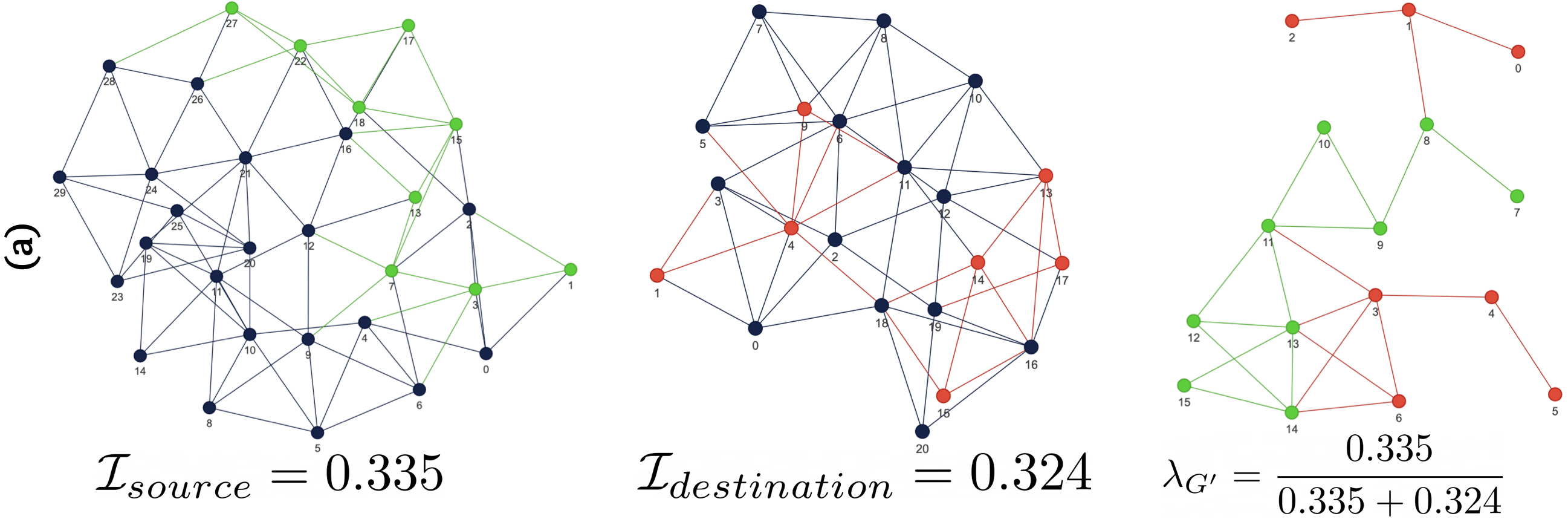}
  \end{minipage}
  
  \vspace{1cm}
  
  \begin{minipage}[b]{0.95\textwidth}
    \includegraphics[width=\columnwidth]{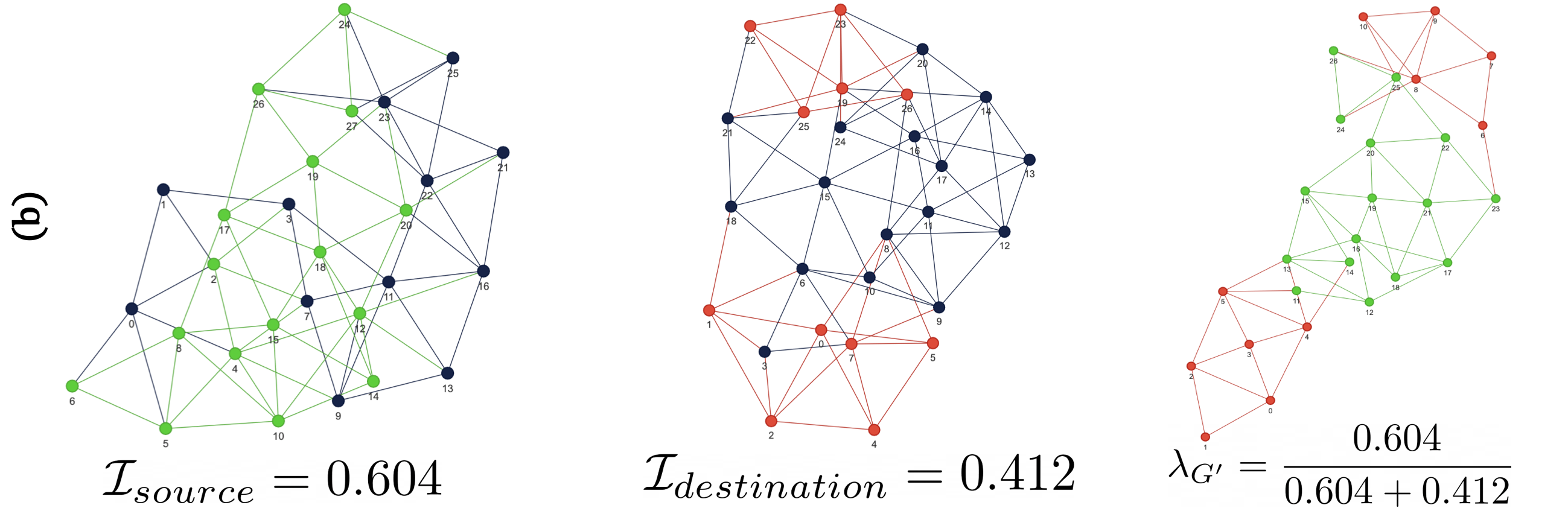}
  \end{minipage}
  \vspace{-0.1in}
  \label{fig:coil-del}
\end{figure*}

\begin{figure*}[h]
  \centering
  \textbf{NCI1}\par\medskip
  \begin{minipage}[b]{0.95\textwidth}
    \includegraphics[width=\columnwidth]{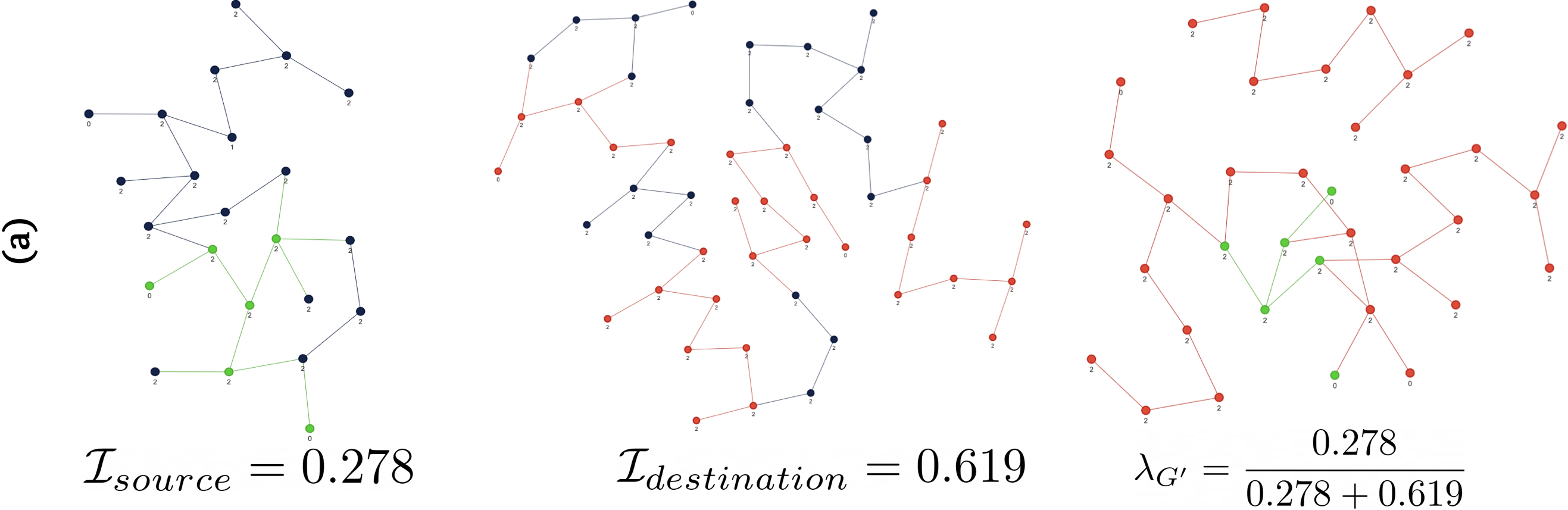}
  \end{minipage}
  
  \vspace{1cm}
  
  \begin{minipage}[b]{0.95\textwidth}
    \includegraphics[width=\columnwidth]{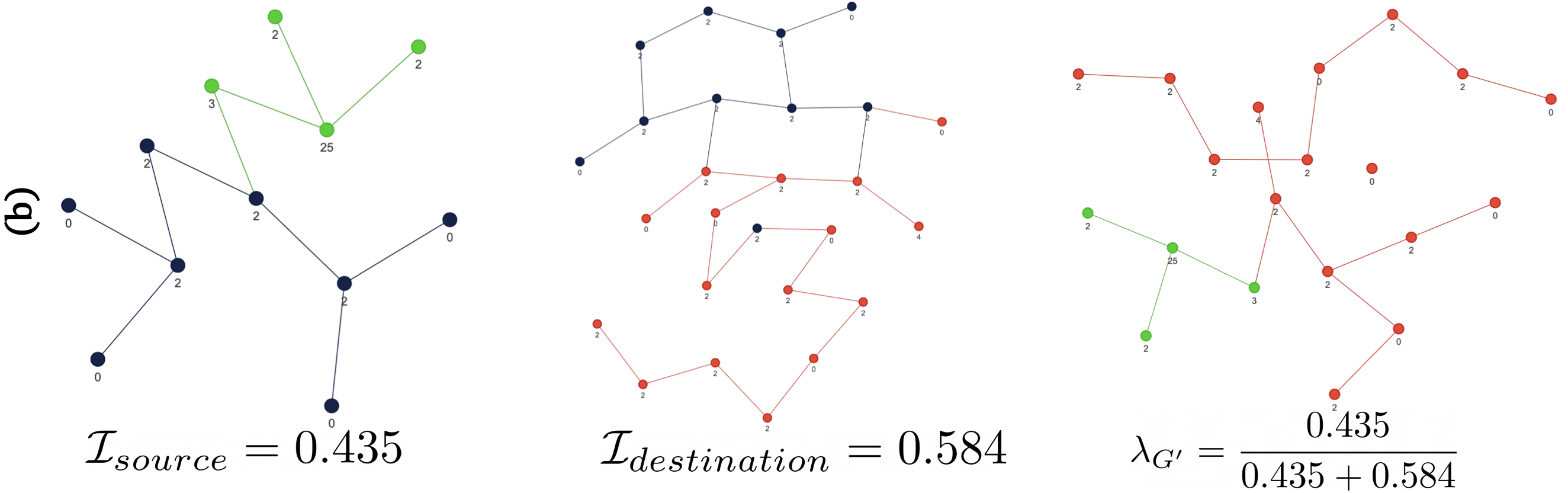}
  \end{minipage}
  \vspace{-0.1in}
  \label{fig:nci}
\end{figure*}

\begin{figure*}[h]
  \centering
  \textbf{COLLAB}\par\medskip
  \begin{minipage}[b]{0.95\textwidth}
    \includegraphics[width=\columnwidth]{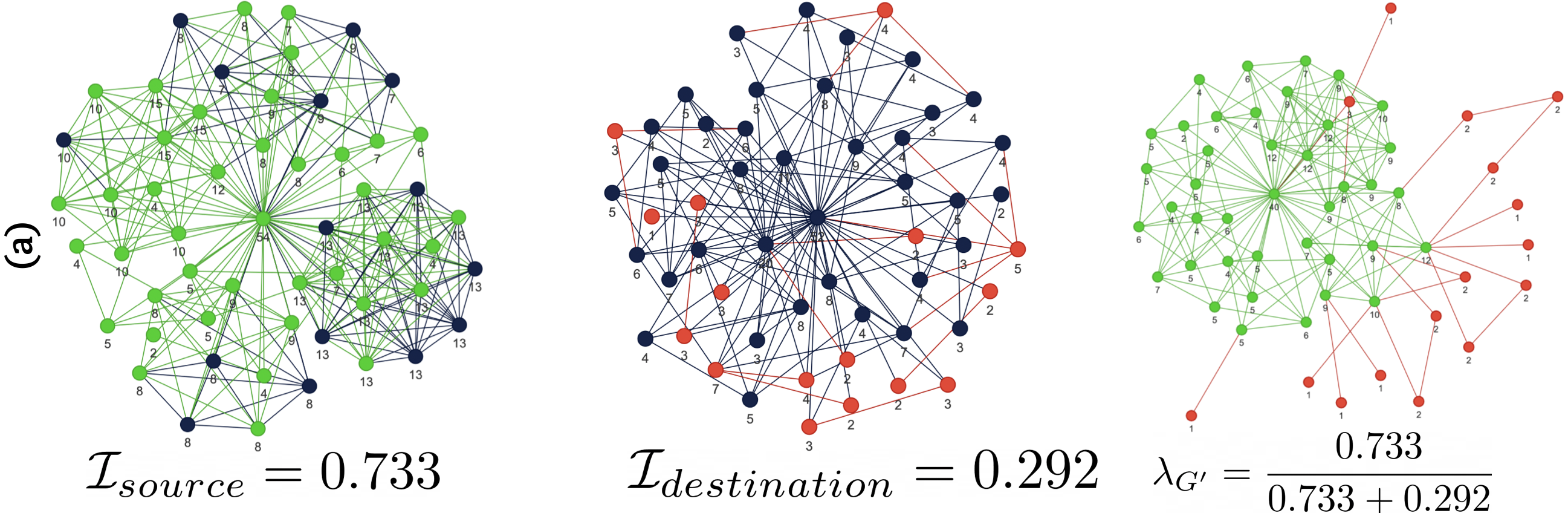}
  \end{minipage}
  
  \vspace{1cm}
  
  \begin{minipage}[b]{0.95\textwidth}
    \includegraphics[width=\columnwidth]{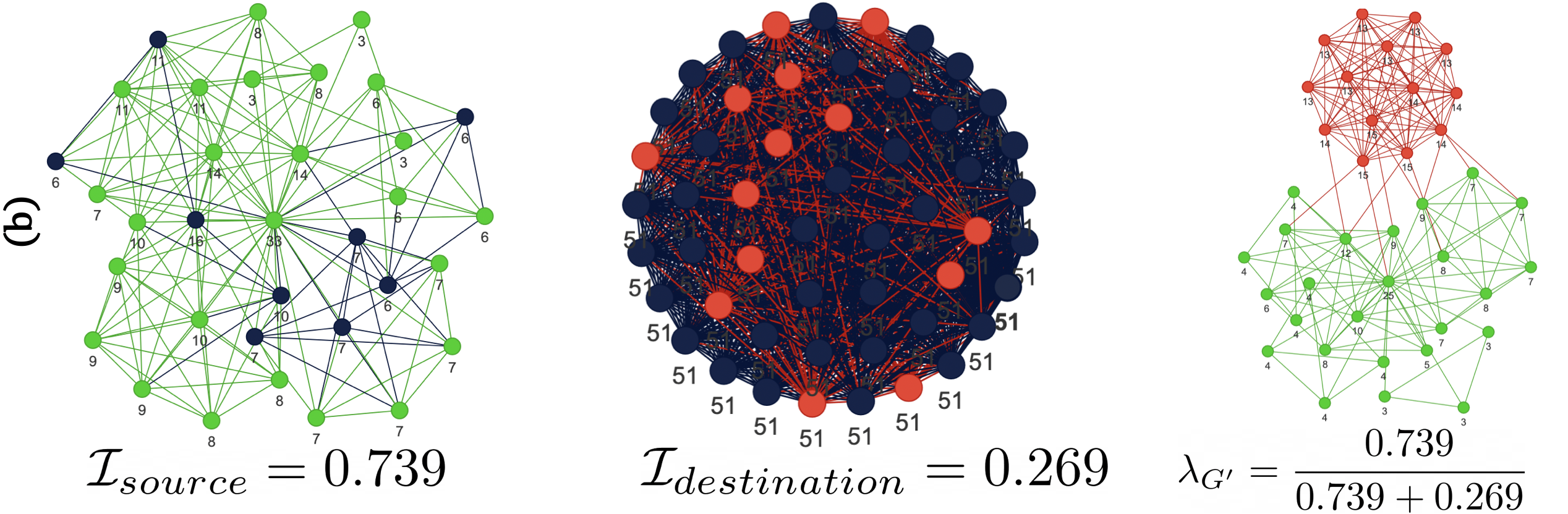}
  \end{minipage}
  \vspace{-0.1in}
  \label{fig:collab}
\end{figure*}

\begin{figure*}[h]
  \centering
  \textbf{Mutagenicity}\par\medskip
  \begin{minipage}[b]{0.95\textwidth}
    \includegraphics[width=\columnwidth]{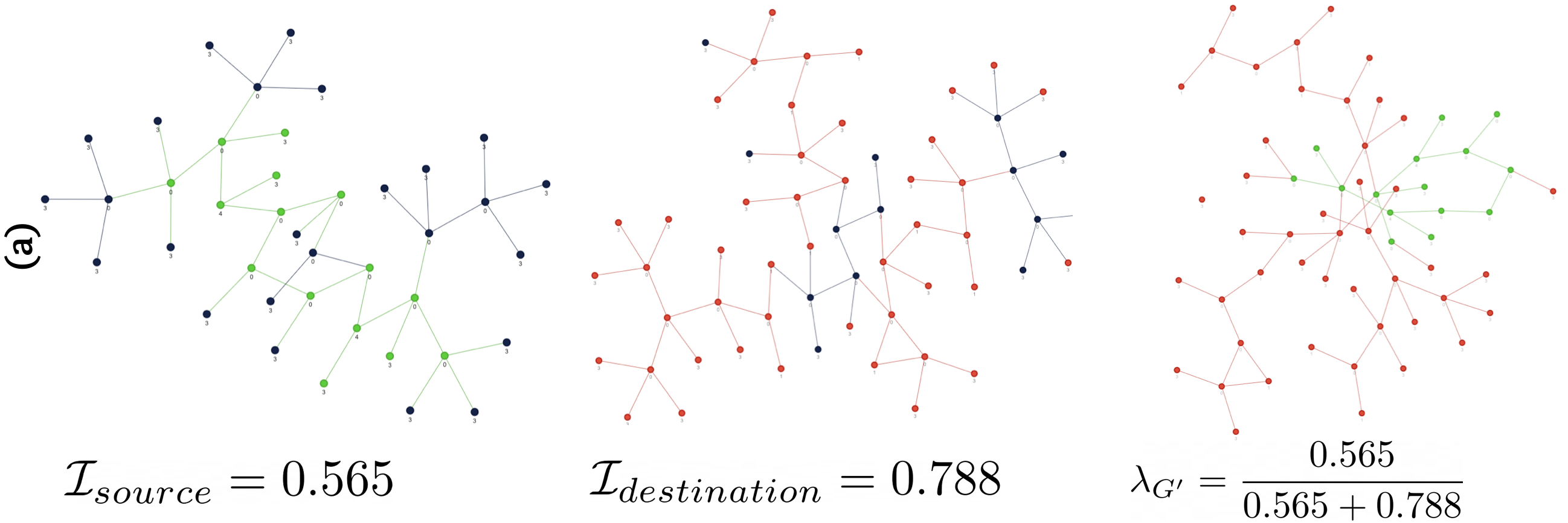}
  \end{minipage}
  
  \vspace{1cm}
  
  \begin{minipage}[b]{0.95\textwidth}
    \includegraphics[width=\columnwidth]{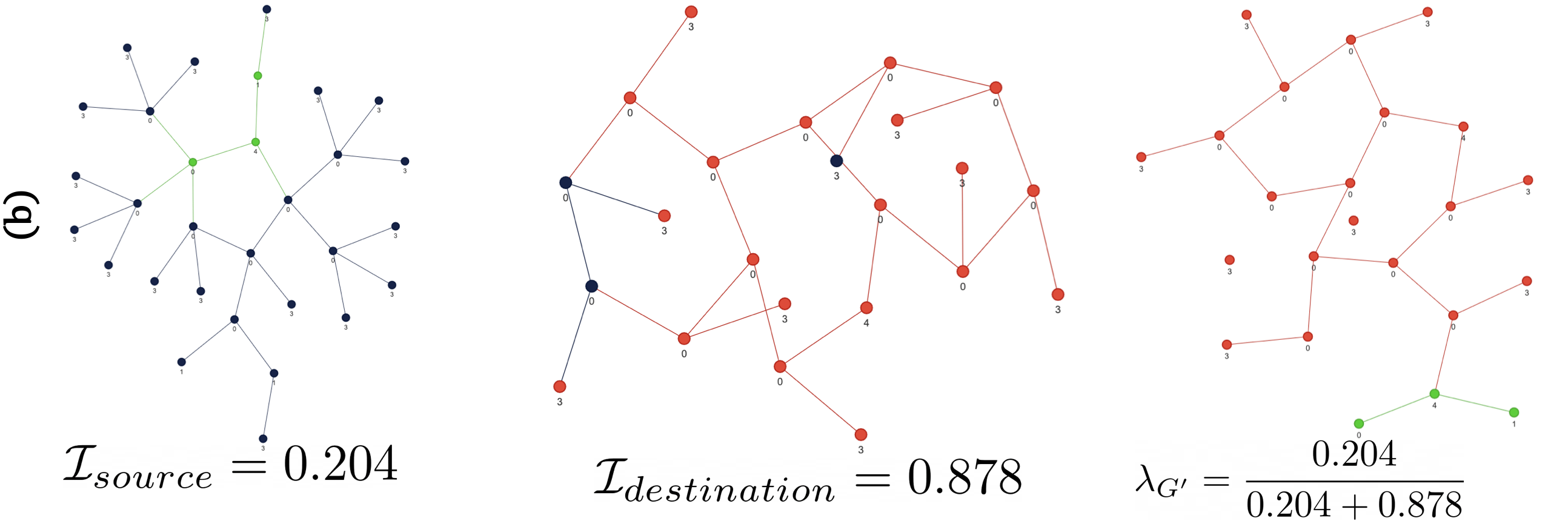}
  \end{minipage}
  \vspace{-0.1in}
  \label{fig:mutagen}
\end{figure*}

\end{document}